\documentclass[letterpaper]{article} 
\usepackage{aaai2026}
\usepackage{times}  
\usepackage{helvet}  
\usepackage{courier}  
\usepackage[hyphens]{url}  
\usepackage{graphicx} 
\usepackage{natbib}  
\usepackage{caption} 
\usepackage{algorithm}
\usepackage{algorithmic}
\usepackage{booktabs}
\usepackage{csvsimple}
\usepackage{tikz}
\usetikzlibrary{positioning}
\usepackage{placeins}

\usepackage{newfloat}
\usepackage{listings}
\DeclareCaptionStyle{ruled}{labelfont=normalfont,labelsep=colon,strut=off} 
\floatstyle{ruled}
\newfloat{listing}{tb}{lst}{}
\floatname{listing}{Listing}
\title{Cost-Sensitive Reversibility of Non-Deterministic Actions}
\author{
    Giuseppe Canonaco\textsuperscript{\rm 1},
    Alberto Pozanco\textsuperscript{\rm 1},
    Daniel Borrajo\textsuperscript{\rm 1}
}
\affiliations{
    \textsuperscript{\rm 1}JPMorgan Chase \& Co. AI Research\\

    giuseppe.canonaco@jpmorgan.com,
    alberto.pozancolancho@jpmorgan.com,
    daniel.borrajo@jpmchase.com
}

\usepackage{bibentry}

\usepackage{amsmath, amsthm, amssymb}
\usepackage{thmtools, thm-restate}

\theoremstyle{definition}
\newtheorem{definition}{Definition}

\newtheorem{remark}{Remark}

\begin{document}

\maketitle

\begin{abstract}
Determining action reversibility requires checking whether the effects of an action can be undone. Recent work has extended reversibility to fully observable non-deterministic (FOND) planning. It distinguishes weak reversibility, where reversal may succeed, from strong-cyclic reversibility, where reversal is guaranteed under fairness. These notions are qualitative: they say whether reversal is possible, but not how costly it is. 

We introduce cost-sensitive reversibility for non-deterministic actions. Given positive action costs, we define weak and strong reversal budgets, $k$-bounded reversibility, and a distinction between finite strong reversal and strong-cyclic reversal with unbounded worst-case cost. We relate the resulting notions in a reversibility lattice, prove cost-aware policy-merging results, and show how to compute restricted weak and strong universal uniform reversal budgets using cost-optimal classical planning and worst-case-optimal FOND search, respectively. We also give an admissible lower-bound test for rejecting $k$-bounded reversibility. These notions support finer-grained safety decisions: a small finite
strong budget bounds recovery cost, whereas a small weak budget
certifies only a cheap possible recovery path. Actions without an
acceptable strong bound may therefore warrant human approval, as in tool-using agent systems.
\end{abstract}

\section{Introduction}

Action reversibility is the problem of deciding whether the effects of an
action can be undone by subsequent actions. It has been studied in
deterministic planning through inverse actions, reverse plans, undoability,
and universal or uniform reversibility notions
~\cite{eiter2008undoing,chrpa2012determining,daum2016practical,
morak2020reversibility,chrpa2021universal,faber2021determining}.
Reversible actions are useful because they identify choices that do not
necessarily commit the agent to an unrecoverable state, which is important
for safe online acting, dead-end avoidance, and plan analysis
~\cite{cserna2018avoiding,chrpa2022speeding}.

In fully observable non-deterministic (FOND) planning, actions may have
several possible outcomes, and solutions are policies rather than fixed
action sequences~\cite{cimatti2003weak}.~\citet{chrpa2024weak} extended
action reversibility to this setting. Their notions distinguish weak
reversibility, where each outcome can be undone along some successful
trajectory, from strong reversibility, where a policy guarantees reversal
under the standard fairness assumption for strong-cyclic solutions
~\cite{aminof2020stochastic}. They also study universal variants, requiring
reversal from every applicable state, and uniform variants, requiring one
common reverse policy.

However, qualitative reversibility is not always a sufficient safety signal.
An action may be reversible in principle, but only after a long, expensive,
or operationally disruptive recovery process. Conversely, an action whose
outcomes can be reversed within a small budget is much less consequential:
even if it has an undesirable outcome, the agent can recover at a bounded
cost. Thus, the relevant question is often not only whether an action is
reversible, but whether it is reversible within an acceptable recovery
budget.

This paper develops such a cost-sensitive view of non-deterministic action
reversibility. Recent work on cost-optimal FOND planning shows how
arbitrary positive action costs can be incorporated into non-deterministic
planning~\cite{aineto2025cost}, enabling quantitative questions about the
cost of acting under non-determinism. We apply this perspective to
reversibility: the cost of interest is not the cost of reaching an ordinary
goal, but the recovery cost incurred after an action has produced one of
its possible outcomes. Accordingly, our work is not a cost-optimal
implementation of the qualitative tests of~\citet{chrpa2024weak}; rather,
costs refine the reversibility notions themselves. The qualitative theory
asks whether a reverse witness exists, whereas we ask whether such a
witness exists within a budget, whether a strong witness has finite
worst-case cost, and how finite-budget reversal differs from strong-cyclic
reversal whose qualitative guarantee may rely on fairness, but whose cost is
unbounded.

We formalize this idea through $k$-bounded reversibility: an action is
reversible within budget $k$ if its outcomes can be undone by a reverse
witness whose relevant cost is at most $k$. For weak reversibility, the
cost is optimistic, taking the cheapest successful reversal trajectory. For
strong reversibility, the cost is pessimistic, taking the worst terminal
trajectory of the reverse policy; cyclic strong witnesses are therefore
separated from finite-budget ones.

Cost-sensitive reversibility also provides a useful abstraction for
tool-using and LLM-based agents. If an action has a certified small strong reversal budget, it can be treated as lower consequence. If it is irreversible, only weakly reversible, or lacks an acceptable finite strong budget, the agent can escalate to a human before acting. This gives a planning-theoretic criterion for autonomy
boundaries based on recoverability. As a simple illustration, a shopping agent may safely add an item to a cart
if this can be undone cheaply, but may need human approval before placing
an order whose reversal requires an expensive return process.

This paper:
\begin{itemize}
    \item  defines weak and strong reversal costs, $k$-bounded
    reversibility, and finite vs. strong-cyclic reversal regimes;
    \item  establishes a lattice relating bounded reversibility, qualitative
    reversibility, strong-cyclic reversibility, and irreversibility;
    \item generalizes qualitative policy-merging results to show how weak and strong reversal budgets compose;
    \item compiles weak reversal budgets to cost-optimal classical
    planning and strong reversal budgets to worst-case-optimal FOND search;
    \item  gives an admissible lower-bound test for rejecting $k$-bounded
    weak reversibility; and
    \item evaluates these certificates on FOND benchmarks under several action-cost models.
\end{itemize}

\section{Preliminaries}

\paragraph{FOND planning with costs.}
We adopt the SAS$^+$/FOND formalism with action costs~\cite{backstrom1995complexity, cimatti2003weak, aineto2025cost}. Let $V$ be a finite
set of state variables, where each variable $v\in V$ has a finite domain
$\mathrm{dom}(v)$. A \emph{fact} is a pair $(v,x)$ with
$x\in\mathrm{dom}(v)$. A \emph{variable assignment} $\Sigma$ over $V$ is a
set of facts that assigns at most one value to each variable. We write
$\mathrm{vars}(\Sigma)=\{v\mid (v,x)\in\Sigma\}$ for the variables assigned
by $\Sigma$. A \emph{state} is a complete variable assignment over $V$, and
$S(V)$ denotes the set of all states. We write $\Sigma\subseteq s$ when the
state $s$ agrees with the partial assignment $\Sigma$.

An action $a$ over $V$ is a pair $a=(\mathrm{pre}(a),\mathrm{eff}(a))$, where $\mathrm{pre}(a)$ is a variable assignment, called the precondition of
$a$, and $\mathrm{eff}(a)$ is a non-empty set of variable assignments, called
the possible effects of $a$. The action $a$ is deterministic iff
$|\mathrm{eff}(a)|=1$. For an effect $e\in\mathrm{eff}(a)$, the
\emph{determinization} of $a$ with respect to $e$ is $a^d_e=(\mathrm{pre}(a),\{e\})$.
Additionally, we write $\mathrm{vars}(a)
  =
  \mathrm{vars}(\mathrm{pre}(a))
  \cup
  \bigcup_{e\in\mathrm{eff}(a)} \mathrm{vars}(e)
$.

An action $a$ is applicable in a state $s$ iff
$\mathrm{pre}(a)\subseteq s$. We write $\alpha(s,A)$ for the set of actions
from $A$ applicable in $s$. If $a^d_e$ is applicable in
$s$, its application is
$
  \gamma(s,a^d_e)
  =
  \{(v,x)\in s\mid v\notin\mathrm{vars}(e)\}\cup e
$.
The application of a non-deterministic action $a$ in $s$ is the set
$
  \delta(s,a)
  =
  \{\gamma(s,a^d_e)\mid e\in\mathrm{eff}(a)\}
$.
For a deterministic action $a^d_e$, we also write
$
  \mathrm{ha}(a^d_e)=\gamma(\mathrm{pre}(a),a^d_e)
$
for the partial assignment that necessarily holds after applying
$a^d_e$ in any state satisfying $\mathrm{pre}(a)$.

A FOND planning task with costs is a tuple
$
  P=\langle V,A,s_0,G,c\rangle
$,
where $A$ is a finite set of actions over $V$, $s_0\in S(V)$ is the initial
state, $G$ is a partial assignment representing the goal, and
$c:A\to\mathbb{N}^{+}$ assigns a strictly positive cost to every action.
We assume the standard fairness assumption for strong cyclic solutions:
if an action is applied infinitely often, then each of its possible effects
occurs infinitely often~\cite{aminof2020stochastic}.

\paragraph{Policies, reachability, and terminals.}
A \emph{policy} $\Pi$ is a binary relation over states and applicable actions;
that is,
$
  \Pi\subseteq \{(s,a)\mid s\in S(V),\ a\in\alpha(s,A)\}
$.
Thus, a policy may relate a state to more than one applicable action. We write
$
  \sigma(\Pi)=\{s\in S(V)\mid \exists a\in A : (s,a)\in\Pi\}
$
for the set of states in which $\Pi$ assigns at least one action.

The states reachable from a state $s$ under $\Pi$, denoted
$\mathrm{Reach}(\Pi,s)$, are the least set $R\subseteq S(V)$ such that
$s\in R$ and, whenever $q\in R$, $(q,a)\in\Pi$, and $q'\in\delta(q,a)$,
then $q'\in R$.

A \emph{$\Pi$-trajectory from $s$} is either a finite alternating sequence
$
  s_0,a_0,s_1,a_1,\ldots,a_{n-1},s_n
$
with $s_0=s$, $(s_i,a_i)\in\Pi$, and
$s_{i+1}\in\delta(s_i,a_i)$ for every $0\leq i<n$, or an infinite
alternating sequence
$
  s_0,a_0,s_1,a_1,\ldots
$
satisfying the same conditions for every $i\geq 0$. A finite trajectory is
\emph{terminal} if its last state $s_n$ satisfies $s_n\notin\sigma(\Pi)$.
The set of terminal states reachable from $s$ under $\Pi$ is denoted
$\tau(\Pi,s)$. Since $S(V)$ is finite, the transition graph induced by
$\Pi$ contains an infinite path from $s$ iff it contains a
cycle reachable from $s$.

\paragraph{Trajectory cost, best case, and worst case.}
The cost of a finite $\Pi$-trajectory
$
  \rho=s_0,a_0,s_1,\ldots,a_{n-1},s_n
$
is
$
  \mathrm{cost}(\rho)=\sum_{i=0}^{n-1} c(a_i)
$.
The empty trajectory has cost $0$.

For a policy $\Pi$ and states $t,s'\in S(V)$, let
$\mathrm{TermTraj}_t(\Pi,s')$ be the set of finite terminal
$\Pi$-trajectories starting in $s'$ and ending in $t$. We define
$$
  \mathrm{best}_{t}(\Pi,s')
  =
  \min_{\rho\in\mathrm{TermTraj}_{t}(\Pi,s')} \mathrm{cost}(\rho),
$$
with the convention that $\mathrm{best}_t(\Pi,s')=+\infty$ if
$\mathrm{TermTraj}_t(\Pi,s')=\emptyset$.

For the worst-case metric, we use the extended value $\bowtie$ to denote
unbounded finite cost. We order finite costs below $\bowtie$, i.e.,
$k<\bowtie$ for every $k\in\mathbb{N}$. We define
$\mathrm{worst}_t(\Pi,s')=\bowtie$ whenever the transition graph induced by
$\Pi$ contains a cycle reachable from $s'$; under fairness, such a cycle
may still be exited eventually, but the number of traversals before exit is
unbounded. Otherwise, the reachable part of the graph is acyclic and
finite, and we define
$$
  \mathrm{worst}_{t}(\Pi,s')
  =
  \max_{\rho\in\mathrm{TermTraj}_{t}(\Pi,s')} \mathrm{cost}(\rho),
$$
again with value $+\infty$ if
$\mathrm{TermTraj}_t(\Pi,s')=\emptyset$. In the sequel,
$\mathrm{worst}_t$ is used only for strong reverse witnesses, for which
terminal trajectories to the target exist from every relevant state; hence
the only non-finite worst-case value that arises for such witnesses is
$\bowtie$.

\paragraph{State-specific reversal.}
Fix an action $a\in A$ to be reversed and a set of states $S\subseteq S(V)$.
We write
$
  S_a=\{s\in S\mid \mathrm{pre}(a)\subseteq s\}
$
for the states in $S$ in which $a$ is applicable.

Let $s\in S_a$. A policy $\Pi$ \emph{weakly reverts} $a$ from $s$ if
$
  s\in\tau(\Pi,s')
  \text{ for every } s'\in\delta(s,a)
$~\cite{chrpa2024weak}.
Thus, after every possible outcome of $a$, there exists at least one finite
$\Pi$-trajectory that returns to the original state $s$.

A policy $\Pi$ \emph{strongly reverts} $a$ from $s$ if, for every
$s'\in\delta(s,a)$,
$
  \tau(\Pi,s')=\{s\}
$,
and every state reachable from $s'$ under $\Pi$ can reach $s$ under $\Pi$~\cite{chrpa2024weak}.
So, $s$ is the only terminal state reachable from every outcome of $a$.

\paragraph{Reverse witnesses and uniformity.}
A \emph{weak $S$-reverse witness} for $a$ is a state-indexed family of
policies
$
  \mathcal{P}=\{\Pi_s\}_{s\in S_a}
$
such that $\Pi_s$ weakly reverts $a$ from $s$ for every $s\in S_a$.
Similarly, a \emph{strong $S$-reverse witness} for $a$ is a family
$\mathcal{P}=\{\Pi_s\}_{s\in S_a}$ such that $\Pi_s$ strongly reverts $a$
from $s$ for every $s\in S_a$.

A reverse witness $\mathcal{P}=\{\Pi_s\}_{s\in S_a}$ is \emph{uniform} if
there exists a policy $\Pi$ such that $\Pi_s=\Pi$ for every $s\in S_a$.
In that case we say that $\Pi$ induces the uniform witness
$\mathcal{P}$. This follows the uniformity distinction used for action reversibility~\cite{chrpa2024weak, chrpa2021universal, morak2020reversibility}.

\paragraph{Qualitative reversibility and irreversibility.}
The action $a$ is \emph{weakly $S$-reversible} if it admits a weak
$S$-reverse witness, and \emph{strongly $S$-reversible} if it admits a
strong $S$-reverse witness. It is \emph{weakly uniformly $S$-reversible},
respectively \emph{strongly uniformly $S$-reversible}, if the corresponding
witness can be chosen uniform.

The action $a$ is \emph{$S$-irreversible} if no relevant state admits even
a weak state-specific reverting policy, that is, if for every $s\in S_a$
there is no policy $\Pi$ such that
$
  s\in\tau(\Pi,s')
  \text{ for every } s'\in\delta(s,a)
$~\cite{chrpa2024weak}.
Universal variants are obtained by taking $S=S(V)$. Uniformity is an
independent requirement: it means that the reverse witness must be induced
by a single policy.

\FloatBarrier

\section{Non-deterministic $k$-bounded Action Reversibility}

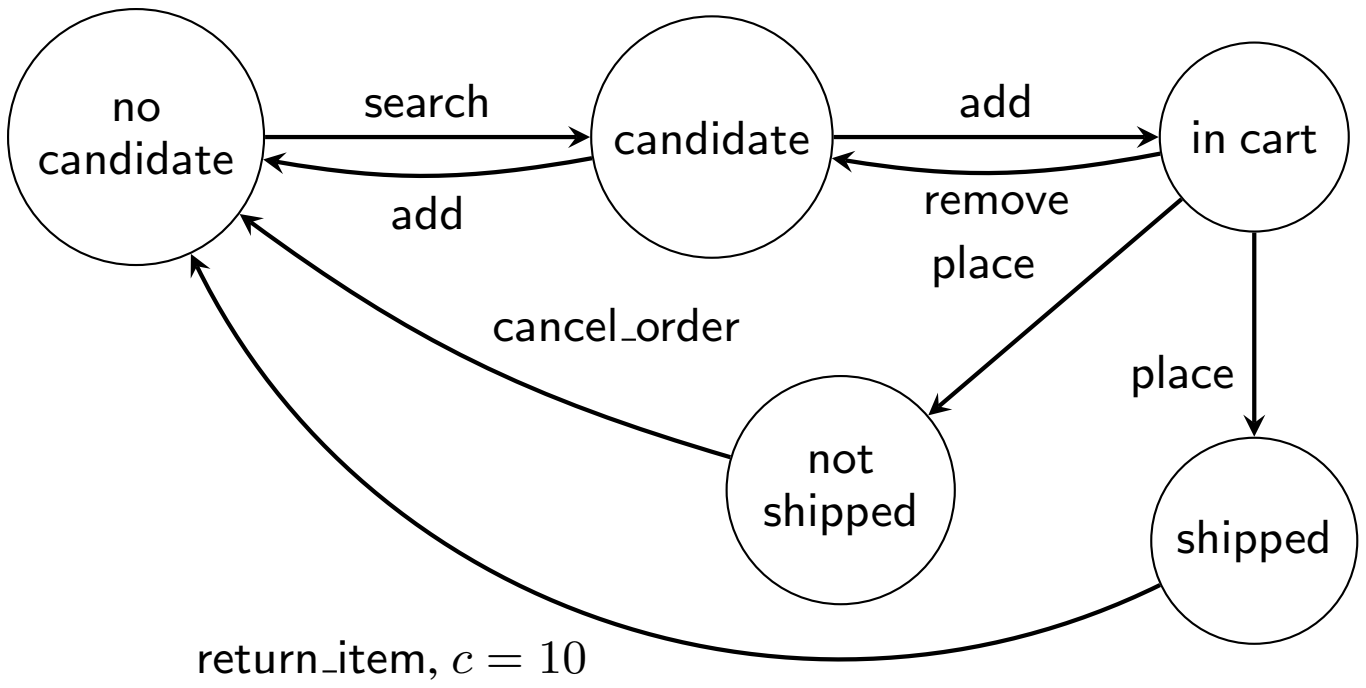
\begin{figure}[t]
\centering
\begin{tikzpicture}[
    >=stealth,
    node distance=2.15cm,
    state/.style={
        circle,
        draw,
        align=center,
        minimum size=1.25cm,
        font=\small
    },
    edge/.style={
        ->,
        thick,
        font=\small
    }
]
\node[state] (no) {$\mathsf{no}$\\ $\mathsf{candidate}$};
\node[state, right=of no] (cand) {$\mathsf{candidate}$};
\node[state, right=of cand] (cart) {$\mathsf{in}~\mathsf{cart}$};
\node[state, below left=1.35cm and 1.75cm of cart] (notship)
    {$\mathsf{not}$\\ $\mathsf{shipped}$};
\node[state, below=1.35cm and 1.75cm of cart] (ship)
    {$\mathsf{shipped}$};

\draw[edge] (no) -- node[above] {$\mathsf{search}$} (cand);

\draw[edge] (cand) -- node[above] {$\mathsf{add}$} (cart);
\draw[edge] (cand) to[bend left=10]
    node[below] {$\mathsf{add}$} (no);

\draw[edge] (cart) to[bend left=10]
    node[below] {$\mathsf{remove}$} (cand);

\draw[edge] (cart) -- node[above left] {$\mathsf{place}$} (notship);
\draw[edge] (cart) -- node[below left] {$\mathsf{place}$} (ship);

\draw[edge] (notship) to[bend left=10]
    node[above right] {$\mathsf{cancel\_order}$} (no);

\draw[edge] (ship) to[bend left=45]
    node[below left] {$\mathsf{return\_item}$, $c=10$} (no);

\end{tikzpicture}
\caption{Transition graph for the online shopping-agent example. All
actions have unit cost except $\mathsf{return\_item}$, which has cost
$10$.}
\label{fig:shopping-agent}
\vspace{-0.2in}
\end{figure}

The notions of weak and strong reversibility above, introduced for non-deterministic actions by~\citet{
chrpa2024weak}, are qualitative: they
record whether an action can be undone, but not at what cost. We refine
them by adding a budget dimension. Reversal is witnessed by a state-indexed
family of policies, one policy for each relevant state in which the action
may be applied. The uniform case is obtained by restricting this family to
be constant. 

The asymmetry between weak and strong budgets follows the qualitative
semantics. Weak reversal requires that, from each relevant outcome, some
successful recovery trajectory exists; hence its budget is the cost of the
cheapest such trajectory. Strong reversal requires a policy to guarantee
return despite all non-deterministic outcomes; hence its budget is the
worst terminal trajectory cost, and a reachable positive-cost cycle makes
this worst-case cost unbounded. Since FOND models do not provide outcome
probabilities, we do not consider expected reversal cost. Otherwise, we would be moving from FOND toward stochastic shortest-path planning~\cite{bertsekas1991analysis}.

To better elucidate the concepts we are about to introduce, we use the shopping-agent graph in Figure~\ref{fig:shopping-agent} as a
running example. The agent acts on behalf of a user who wants to buy an item
satisfying constraints such as product type, price, delivery date, and
seller requirements. States abstract the purchasing status. The agent can
search for a candidate item, add it to the cart, remove it, place an order,
cancel an order that has not shipped, or return an item that has shipped.
All actions have unit cost except $\mathsf{return\_item}$, which has cost
$10$. The action $\mathsf{add}$ may fail because the product becomes
unavailable before the add-to-cart step completes.

We now formalize the cost notions and instantiate them on this example.

\subsection{Worst- and best-case reversal cost}

\begin{definition}[Worst- and best-case reversal cost]
\label{def:wrc-brc-family}
Let $a\in A$, $S\subseteq S(V)$, and
$S_a=\{s\in S\mid \mathrm{pre}(a)\subseteq s\}$. Let
$
  \mathcal{P}=\{\Pi_s\}_{s\in S_a}
$
be a weak or strong $S$-reverse witness for $a$. The
\emph{worst-case reversal cost} of $\mathcal{P}$ over $S$ is
$$
  \mathrm{wrc}_S(\mathcal{P},a)
  =
  \max_{\substack{s\in S_a\\ s'\in\delta(s,a)}}
  \mathrm{worst}_s(\Pi_s,s').
$$
The \emph{best-case reversal cost} of $\mathcal{P}$ over $S$ is
$$
  \mathrm{brc}_S(\mathcal{P},a)
  =
  \max_{\substack{s\in S_a\\ s'\in\delta(s,a)}}
  \mathrm{best}_s(\Pi_s,s').
$$
If $\mathcal{P}$ is uniform and induced by a policy $\Pi$, we may write
$\mathrm{wrc}_S(\Pi,a)$ and $\mathrm{brc}_S(\Pi,a)$ as shorthand for the
costs of the induced constant family.
\end{definition}

The outer maximum in $\mathrm{brc}_S$ is not an adversarial choice during
execution. Rather, it aggregates the cheapest successful reversal cost over
all relevant pre-states $s\in S_a$ and all possible outcomes
$s'\in\delta(s,a)$. Thus a finite bound on $\mathrm{brc}_S$ requires that
every relevant outcome have some successful reversal trajectory within the
bound. By contrast, $\mathrm{wrc}_S$ bounds the worst reversal trajectory
and is therefore sensitive to reachable cycles.

\paragraph{Running example.}
Consider reversing $\mathsf{place}$ from $\mathsf{in\_cart}$ in
Figure~\ref{fig:shopping-agent}. A natural reverse policy is
\begin{align*}
  \Pi_{\mathsf{place}}
  =
  \{
    (\mathsf{not\_shipped},\mathsf{cancel\_order}),\\
    (\mathsf{shipped},\mathsf{return\_item}),
    (\mathsf{no\_candidate},\mathsf{search}),\\
    (\mathsf{candidate},\mathsf{add})
  \}.
\end{align*}
From $\mathsf{not\_shipped}$, a successful trajectory back to
$\mathsf{in\_cart}$ has cost $1+1+1=3$. From $\mathsf{shipped}$, a
successful trajectory has cost $10+1+1=12$. Thus the best-case reversal
cost of this witness is $12$. However, because $\mathsf{add}$ may return
from $\mathsf{candidate}$ to $\mathsf{no\_candidate}$, the same policy has
a reachable cycle during recovery, and its strong worst-case cost is
$\bowtie$.

\subsection{$k$-bounded reversibility}

\begin{definition}[$k$-bounded reversibility]
\label{def:kbounded}
Let $k\in\mathbb{N}$ and $S\subseteq S(V)$. An action $a$ is:
\begin{itemize}
  \item \emph{$k$-bounded weakly $S$-reversible} if there exists a weak
  $S$-reverse witness $\mathcal{P}$ for $a$ such that
  $
    \mathrm{brc}_S(\mathcal{P},a)\le k;
  $
  \item \emph{$k$-bounded strongly $S$-reversible} if there exists a strong
  $S$-reverse witness $\mathcal{P}$ for $a$ such that
  $
    \mathrm{wrc}_S(\mathcal{P},a)\le k.
  $
\end{itemize}
The action is \emph{$k$-bounded weakly (respectively strongly) uniformly $S$-reversible}, if the
corresponding witness can be chosen uniformly. Such a witness is called a
\emph{$k$-bounded weak (uniform)}, respectively \emph{$k$-bounded strong (uniform)},
\emph{$S$-reverse witness}.
\end{definition}

Universal variants are obtained by taking $S=S(V)$. Universality and
uniformity are independent axes: universality quantifies over all states in
which $a$ is applicable, whereas uniformity requires the state-indexed
witness to be induced by a single policy.

\paragraph{Running example.}
The action $\mathsf{add}$, applied in $\mathsf{candidate}$, has outcomes
$\mathsf{in\_cart}$ and $\mathsf{no\_candidate}$. The policy
$$
  \Pi_{\mathsf{add}}
  =
  \{
    (\mathsf{in\_cart},\mathsf{remove}),
    (\mathsf{no\_candidate},\mathsf{search})
  \}
$$
returns deterministically to $\mathsf{candidate}$ from both outcomes.
Hence $\mathsf{add}$ is $1$-bounded strongly reversible, and therefore also
$1$-bounded weakly reversible. By contrast, the action $\mathsf{place}$ is
not weakly $k$-bounded for any $k<12$ in this graph: the
$\mathsf{shipped}$ outcome requires $\mathsf{return\_item}$, then
$\mathsf{search}$, and then a successful $\mathsf{add}$ to recover
$\mathsf{in\_cart}$.

\subsection{Acyclic, cyclic, and the cost stratification}

A strong reverse witness need not be acyclic. This is because we adopt the
strong-cyclic interpretation of FOND policies: under the fairness assumption,
a policy may contain reachable cycles and still guarantee eventual return to
the target state. Fairness rules out executions that remain forever in a
cycle only because some enabled outcome is ignored indefinitely. It does
not, however, impose a uniform bound on how many times a cycle may be
traversed before progress occurs. Hence, with strictly positive action
costs, a reachable cycle yields arbitrarily expensive finite reversal
trajectories, even though reversal is guaranteed qualitatively. This
distinction, invisible in the qualitative theory, is central once costs are
introduced.

\begin{definition}[Cost regimes of strong reverse witnesses]
\label{def:regimes}
Let $\mathcal{P}=\{\Pi_s\}_{s\in S_a}$ be a strong $S$-reverse witness for
$a$.

For a fixed $s\in S_a$, the component $\Pi_s$ is \emph{bounded for $s$} if,
for every $s'\in\delta(s,a)$, the transition graph induced by $\Pi_s$
contains no cycle reachable from $s'$. Otherwise, $\Pi_s$ is
\emph{cyclic for $s$}.

The witness $\mathcal{P}$ is:
\begin{itemize}
  \item \emph{bounded} if every component $\Pi_s$ is bounded for its
  corresponding state $s$;
  \item \emph{strong cyclic} if some component $\Pi_s$ is cyclic for its
  corresponding state $s$.
\end{itemize}
\end{definition}

\begin{restatable}[Acyclicity and finite worst-case cost]
  {proposition}{acyclicFinite}
\label{prop:acyclic-finite}
Let $\mathcal{P}$ be a strong $S$-reverse witness for $a$ and let
$c:A\to\mathbb{N}^{+}$. Then
$
  \mathrm{wrc}_S(\mathcal{P},a)\in\mathbb{N}
$
if and only if $\mathcal{P}$ is bounded.
\end{restatable}

The proof of Proposition~\ref{prop:acyclic-finite} is in Appendix~\ref{appx:propproofs}.

\paragraph{Running example.}
The policy $\Pi_{\mathsf{place}}$ above is strong-cyclic. If recovery
reaches $\mathsf{candidate}$, the action $\mathsf{add}$ may either return
to the target $\mathsf{in\_cart}$ or lose the candidate and return to
$\mathsf{no\_candidate}$. Under fairness, repeated attempts eventually
produce the $\mathsf{in\_cart}$ outcome, so the policy is qualitatively
strong. Since the loop through $\mathsf{no\_candidate}$ and
$\mathsf{candidate}$ may be traversed arbitrarily many times before
success, and all costs are positive, the worst-case reversal cost is
$\bowtie$.

\subsection{The reversibility lattice}

We now relate the qualitative notions to their cost-bounded counterparts.
Because uniform reversibility is obtained by restricting witnesses to be
constant families, we state the lattice once for both modes.

Let
$
  \mu\in\{\mathsf{nu},\mathsf{u}\}
$,
where $\mathsf{nu}$ denotes the non-uniform mode and $\mathsf{u}$ denotes
the uniform mode. A witness is \emph{admissible in mode $\mathsf{nu}$} if
it is any state-indexed reverse witness, and \emph{admissible in mode
$\mathsf{u}$} if it is uniform.

We use the following predicates:
\begin{itemize}
    \item $\mathsf{WkRev}^{\mu}(a,S)$
    : $a$ admits a weak $S$-reverse witness admissible in mode $\mu$,
    \item $\mathsf{StrRev}^{\mu}(a,S)$
    : $a$ admits a strong $S$-reverse witness admissible in mode $\mu$,
    \item $\mathsf{Wk}^{\mu}_k(a,S)$
    : $a$ admits such a weak witness with $\mathrm{brc}_S\le k$,
    \item $\mathsf{Str}^{\mu}_k(a,S)$
    : $a$ admits such a strong witness with $\mathrm{wrc}_S\le k$.
\end{itemize}

Irreversibility is independent of the mode:
$
  \mathsf{Irr}(a,S)
$
means that no relevant state admits even a weak state-specific reverting
policy.

\begin{definition}[Bounded and cyclic strong reversibility]
\label{def:bounded-predicates}
For $\mu\in\{\mathsf{nu},\mathsf{u}\}$:
\begin{itemize}
  \item $\mathsf{BStr}^{\mu}(a,S)$ holds iff $a$ admits a bounded strong
  $S$-reverse witness admissible in mode $\mu$;
  \item $\mathsf{SCyc}^{\mu}(a,S)$ holds iff $a$ is strongly
  $S$-reversible in mode $\mu$, but admits no bounded strong
  $S$-reverse witness admissible in mode $\mu$.
\end{itemize}
Equivalently,
$$
  \mathsf{SCyc}^{\mu}(a,S)
  \quad\text{iff}\quad
  \mathsf{StrRev}^{\mu}(a,S)\ \text{and not}\ \mathsf{BStr}^{\mu}(a,S).
$$
We also define
$$
  \mathsf{BWk}^{\mu}(a,S)
  \quad\text{iff}\quad
  \exists k\in\mathbb{N}:\mathsf{Wk}^{\mu}_k(a,S).
$$
Finally, when it exists, the exact strong worst-case budget in mode $\mu$
is
$$
  k^{\star,\mu}_{\mathrm{str}}(a,S)
  =
  \min\{k\in\mathbb{N}\mid \mathsf{Str}^{\mu}_k(a,S)\}.
$$
Similarly, the exact weak best-case budget in mode $\mu$ is
$$
  k^{\star,\mu}_{\mathrm{wk}}(a,S)
  =
  \min\{k\in\mathbb{N}\mid \mathsf{Wk}^{\mu}_k(a,S)\}.
$$
\end{definition}

\begin{restatable}[Refined reversibility lattice]
  {proposition}{refinedLattice}
\label{prop:refined-lattice}
For every action $a$, state set $S$, mode
$\mu\in\{\mathsf{nu},\mathsf{u}\}$, and all $k\le k'$ in $\mathbb{N}$:
\begin{enumerate}
  \item \emph{Budget monotonicity.}
  $$
    \mathsf{Str}^{\mu}_k(a,S)
    \Rightarrow
    \mathsf{Str}^{\mu}_{k'}(a,S),
    \qquad
    \mathsf{Wk}^{\mu}_k(a,S)
    \Rightarrow
    \mathsf{Wk}^{\mu}_{k'}(a,S).
  $$

  \item \emph{Strong within budget implies weak within budget.}
  $$
    \mathsf{Str}^{\mu}_k(a,S)
    \Rightarrow
    \mathsf{Wk}^{\mu}_k(a,S).
  $$

  \item \emph{Cost dichotomy of strong reversibility.}
  $$
    \mathsf{StrRev}^{\mu}(a,S)
    \iff
    \mathsf{BStr}^{\mu}(a,S)\ \veebar\ \mathsf{SCyc}^{\mu}(a,S),
  $$
  where $\veebar$ denotes exclusive disjunction.

  \item \emph{No cyclic tier for weak reversibility.}
  $$
    \mathsf{BWk}^{\mu}(a,S)
    \iff
    \mathsf{WkRev}^{\mu}(a,S).
  $$

  \item \emph{Qualitative spine.}
  \begin{align*}
    \mathsf{BStr}^{\mu}(a,S)
    &\Rightarrow
    \mathsf{StrRev}^{\mu}(a,S)
    \Rightarrow \\
    &\mathsf{WkRev}^{\mu}(a,S)
    \Rightarrow
    \neg\,\mathsf{Irr}(a,S).
  \end{align*}
  The last implication is not reversible in general.
\end{enumerate}
\end{restatable}

The proof of Proposition~\ref{prop:refined-lattice} is in Appendix~\ref{appx:propproofs}.

\begin{remark}[The strong-cyclic stratum]
The predicate $\mathsf{SCyc}^{\mu}(a,S)$ identifies actions that are
strongly reversible in mode $\mu$ but never with a finite worst-case budget
in that mode. Qualitatively, such actions are strongly reversible: under
fairness, reversal is guaranteed. Quantitatively, however, every admissible
strong witness in mode $\mu$ contains a reachable cycle on some relevant
reversal, and therefore admits terminal trajectories of arbitrarily large
finite cost. This creates a genuine cost stratum that is invisible in the
qualitative theory.
\end{remark}

\section{Merging Reverse Policies under Costs}

The lattice of the previous section classifies reverse witnesses. In
practice, however, reverse policies are often constructed piecewise---for
example, one policy per determinization or one policy per region of the
relevant state space---and then combined. This section studies how the
budget axis behaves under such combinations.

We focus on the case where each piece is a uniform reverse policy over its
own state set. Equivalently, each piece induces a constant reverse witness
over that region. The weak and strong cases behave differently. In the weak
case, adding more actions can only enlarge the
set of successful target-reaching trajectories; hence merging cannot
increase the best-case reversal cost. In the strong case, every outcome
must be handled safely, so we impose an additional disjointness condition
ensuring that each reversal remains confined to the policy that owns its
target. Under that confinement, the worst-case cost of the merge is
inherited from the component policies.

Throughout, $a$ is the action to be reversed. For $i\in\{1,2\}$, let
$S_i\subseteq S(V)$, and write
$
  S_i^a=\{s\in S_i\mid \mathrm{pre}(a)\subseteq s\}
$
for the relevant states of $S_i$. Let $\Pi_i$ be a reverse policy for
$a$ over $S_i$, i.e., a policy inducing a uniform reverse witness over
$S_i$.

We use the following non-interference condition:
$$
\textup{(C1)}\quad
\sigma(\Pi_2)\cap S_1^a=\emptyset
\quad\text{and}\quad
\sigma(\Pi_1)\cap S_2^a=\emptyset .
$$
Thus neither policy assigns an action to a relevant target state of the
other policy.

For strong merging we additionally use the domain-disjointness condition:
$$
\textup{(C2)}\quad
\sigma(\Pi_1)\cap\sigma(\Pi_2)=\emptyset .
$$
Since policies are relations, $\Pi_1\cup\Pi_2$ is a policy even without
\textup{(C2)}. The role of \textup{(C2)} is
confinement: it prevents a trajectory owned by one component from entering
a state where the other component also assigns actions.

\subsection{Weak merging}

For weak reversal, qualitative merging requires only the non-interference
condition \textup{(C1)}~\cite{chrpa2024weak}. The reason is existential: to show weak reversal
from a relevant outcome, it suffices that at least one successful trajectory
to the target remains available after merging. Since policies are
relations, taking the union preserves all transitions of each component and we only need to be careful to preserve terminal states.

\begin{restatable}[Weak merging with costs]
  {theorem}{weakMergingCosts}
\label{thm:weak_merging}
Let $S_1,S_2\subseteq S(V)$. For $i\in\{1,2\}$, let $\Pi_i$ be a
$k_i$-bounded weak uniform $S_i$-reverse policy for $a$. If
\textup{(C1)} holds, then $\Pi_1\cup\Pi_2$ is a weak uniform
$(S_1\cup S_2)$-reverse policy for $a$, and
$
  \mathrm{brc}_{S_1\cup S_2}(\Pi_1\cup\Pi_2,a)
  \le
  \max(k_1,k_2)
$.
No domain-disjointness condition \textup{(C2)} is required. The bound need
not be tight.
\end{restatable}

The proof of Theorem~\ref{thm:weak_merging} is in Appendix~\ref{appx:thmproofs}.

The inequality can be strict. Since \textup{(C2)} is not required,
$\Pi_2$ may assign actions inside the reachable region of $\Pi_1$, or
vice versa. Thus the merged policy may contain new successful
target-reaching trajectories that are cheaper than the successful
trajectories available in either component alone.

The weak merging result extends by induction to any finite family
whose policies pairwise satisfy (C1): their union is a weak uniform
reverse policy with budget at most the maximum component budget.
The formal statement and proof are in Appendix~\ref{appx:thmproofs}.

\subsection{Strong merging}

For strong reversal, merging is more delicate. Although policies are still
relations, the guarantee must hold for every outcome of every action chosen
by the policy. We therefore impose both \textup{(C1)} and \textup{(C2)}~\cite{chrpa2024weak}.
Condition \textup{(C1)} ensures that the target state of one component is
not made non-terminal by the other. Condition \textup{(C2)} ensures that
reachable non-terminal states of one component are not assigned additional
actions by the other. Together, these conditions confine each reversal to
the component policy that owns its target.

\begin{restatable}[Strong merging with cost regimes]
  {theorem}{strongMergingCosts}
\label{thm:strong_merging}
Let $S_1,S_2\subseteq S(V)$. For $i\in\{1,2\}$, let $\Pi_i$ be a
strong uniform $S_i$-reverse policy for $a$. If \textup{(C1)} and
\textup{(C2)} hold, then $\Pi_1\cup\Pi_2$ is a strong uniform
$(S_1\cup S_2)$-reverse policy for $a$. Moreover:
\begin{enumerate}
  \item \emph{Both components bounded.} If each $\Pi_i$ is bounded, then
  $\Pi_1\cup\Pi_2$ is bounded and
  $
    \mathrm{wrc}_{S_1\cup S_2}(\Pi_1\cup\Pi_2,a)
    =
    \max\big(
      \mathrm{wrc}_{S_1}(\Pi_1,a),
      \mathrm{wrc}_{S_2}(\Pi_2,a)
    \big)
  $.
  In particular, if $\Pi_i$ is $k_i$-bounded for $i=1,2$, then
  $
    \mathrm{wrc}_{S_1\cup S_2}(\Pi_1\cup\Pi_2,a)
    \le
    \max(k_1,k_2)
  $.

  \item \emph{One bounded component and one strong-cyclic component.} If
  one component is bounded and the other is strong cyclic, then
  $\Pi_1\cup\Pi_2$ is strong cyclic and
  $
    \mathrm{wrc}_{S_1\cup S_2}(\Pi_1\cup\Pi_2,a)=\ \bowtie
  $.
  The bounded component retains its finite bound on its own relevant
  states.

  \item \emph{Both components strong cyclic.} If both components are strong
  cyclic, then $\Pi_1\cup\Pi_2$ is strong cyclic and
  $
    \mathrm{wrc}_{S_1\cup S_2}(\Pi_1\cup\Pi_2,a)=\ \bowtie
  $.
\end{enumerate}
\end{restatable}

The proof of Theorem~\ref{thm:strong_merging} is in Appendix~\ref{appx:thmproofs}.

The strong merging result likewise extends by induction to any finite
family satisfying pairwise (C1) and (C2). Its union has a finite
worst-case budget exactly when all components are bounded; otherwise
the constructed witness has cost $\bowtie$. The formal statement and
proof are in Appendix~\ref{appx:thmproofs}.

\section{Computing Reversal Budgets by Compilation}
We compute reversal budgets for the universal uniform case using the
restricted variable and action space of~\citet{chrpa2024weak}. Finite
budgets obtained in this space are exact for the restricted compilation
but provide only upper bounds on the corresponding full-domain optima:
an unrestricted reverse policy may be cheaper. Likewise, cyclicity in
the restricted space does not establish that every full-domain strong
reverse policy is cyclic.

\subsection{Shaping the search space}
\label{sub:restrict}

Before compiling, we ask which actions and variables the reverse task must
range over. Two results of~\citet{chrpa2024weak} carry over. The first is a
cost-invariant obstruction to strong universal uniform reversibility. The
second is a sound, but generally one-sided, restriction of the search space.

The structural obstruction of Lemma~9 of~\citet{chrpa2024weak} is
cost-invariant: if an effect of $a$ modifies a variable absent from
$\mathrm{pre}(a)$, then distinct pre-states may collapse to a common
successor, and no strong universal uniform reverse policy can separate
them. Costs cannot repair this obstruction.

\begin{restatable}[Cost-invariant precheck]
  {corollary}{costInvariantPrecheck}
\label{cor:precheck}
Let $D=\langle V,A\rangle$ with $|\mathrm{dom}(v)|\ge 2$ for all
$v\in V$, and let $a\in A$. If
$$
  \bigcup_{e\in\mathrm{eff}(a)}\mathrm{vars}(e)
  \not\subseteq
  \mathrm{vars}(\mathrm{pre}(a)),
$$
then $a$ is not strongly universally uniformly reversible. Hence no finite
strong universal uniform budget exists, and the strong-cyclic regime is
impossible as well, since it presupposes qualitative strong universal
uniform reversibility.
\end{restatable}

When Corollary~\ref{cor:precheck} (proof in Appendix~\ref{appx:thmproofs}) does not hold, we may
search in the restricted domain over the variables in
$\mathrm{pre}(a)$. Let
$$
  A_{\mathrm{pre}}
  =
  \{a'\in A\mid
    \mathrm{vars}(a')\subseteq \mathrm{vars}(\mathrm{pre}(a))\}.
$$
The following theorem is the cost-aware form of Theorem~10 of
\citet{chrpa2024weak}. It certifies universality, but only gives an upper
bound on the full-domain optimal budget.

\begin{restatable}[Cost-aware restriction]
  {theorem}{costRestrict}
\label{thm:restrict}
Let $a\in A$ satisfy
$
  \bigcup_{e\in\mathrm{eff}(a)}\mathrm{vars}(e)
  \subseteq
  \mathrm{vars}(\mathrm{pre}(a)).
$
Suppose that, in the restricted domain
$
  \langle \mathrm{vars}(\mathrm{pre}(a)), A_{\mathrm{pre}}\rangle
$,
there exists a weak, respectively strong, uniform reverse policy for $a$
with budget $k$. Then this policy induces a weak, respectively strong,
universal uniform reverse witness for $a$ in the full domain. Moreover,
$
  k^{\star,\mathsf{u}}_{\mathrm{wk,full}}(a,S(V))\le k
$
in the weak case, and
$
  k^{\star,\mathsf{u}}_{\mathrm{str,full}}(a,S(V))\le k
$
in the strong case.
\end{restatable}

Thus the restricted search returns an exact budget for the restricted
compilation and a sound upper bound for the full problem. The inequality may
be strict: an unrestricted optimal reverse policy may use cheaper detours
through variables outside $\mathrm{vars}(\mathrm{pre}(a))$, which the
restricted compilation forbids. The proof of Theorem~\ref{thm:restrict} is in Appendix~\ref{appx:thmproofs}.

\subsection{Weak budgets by cost-optimal classical planning}
\label{sub:weakcomp}

For weak reversibility, only existence of a successful reversal trajectory
matters. Hence non-deterministic actions can be replaced by their
determinizations, and the relevant budget is the cost of the cheapest
successful reversal for each outcome of $a$.

\begin{restatable}[Restricted weak universal uniform budget]
  {theorem}{weakComp}
\label{thm:weakcomp}
Let $a\in A$ satisfy the precondition of Theorem~\ref{thm:restrict}. For
each $e\in\mathrm{eff}(a)$, consider the classical planning task
$
T^{d}_e
=
\Big\langle
  \big\langle
    \mathrm{vars}(\mathrm{pre}(a)),
    \{(a')^d_{e'} \mid
      a'\in A,\ e'\in\mathrm{eff}(a'),\
      \mathrm{vars}((a')^d_{e'})
      \subseteq \mathrm{vars}(\mathrm{pre}(a))
    \}
  \big\rangle,
  \mathrm{ha}(a^d_e),
  \mathrm{pre}(a)
\Big\rangle .
$
Let $\pi_e$ be a cost-optimal goal plan for $T^d_e$, with
$c(\pi_e)=+\infty$ if $T^d_e$ is unsolvable. If every $T^d_e$ is
solvable, then $a$ is weakly universally uniformly reversible. Moreover,
the induced restricted weak universal uniform policy $\Pi=\bigcup_e\Pi_e$
satisfies
$
  \mathrm{brc}_{S(V)}(\Pi,a)
  =
  \max_{e\in\mathrm{eff}(a)} c(\pi_e).
$
Consequently,
$
  k^{\star,\mathsf{u}}_{\mathrm{wk,res}}(a)
  =
  \max_{e\in\mathrm{eff}(a)} c(\pi_e),
$
and
$
  k^{\star,\mathsf{u}}_{\mathrm{wk,full}}(a,S(V))
  \le
  k^{\star,\mathsf{u}}_{\mathrm{wk,res}}(a).
$
\end{restatable}

Each $T^d_e$ is an ordinary cost-optimal classical planning task. The proof of the above theorem is in Appendix~\ref{appx:thmproofs}.

\subsection{Strong budgets and regimes by worst-case-optimal FOND search}
\label{sub:strongcomp}

Strong reversibility must guarantee return to the target under every
outcome of every action used by the reverse policy. Therefore the reverse
task remains a FOND task. Under costs, we must distinguish three cases for
each restricted reverse task: no strong policy exists; an acyclic strong
policy exists, yielding a finite worst-case budget; or strong policies exist
but all are cyclic, yielding the $\bowtie$ regime in the restricted task.

For each $e\in\mathrm{eff}(a)$, define the restricted FOND reverse task
$$
R_e
=
\Big\langle
  \big\langle
    \mathrm{vars}(\mathrm{pre}(a)),
    A_{\mathrm{pre}}
  \big\rangle,
  \mathrm{ha}(a^d_e),
  \mathrm{pre}(a)
\Big\rangle 
$$
(solved using our cost-aware extension of BOAND*~\cite{aineto2025cost}; see Appendix~\ref{appx:exp_details}).
The per-effect tasks are solved independently, while uniform strong
reversibility is a global property of the family of selected policies.
Therefore, the results below are certificates for the selected family:
when the selected per-effect policies are pairwise disjoint, they can be
merged and the resulting budget is exact for that constructed restricted
witness. If the selected policies are not mergeable, this is inconclusive:
there may exist other optimal, or even non-optimal but finite, per-effect
policies that do merge.

\begin{restatable}[Restricted strong universal uniform budget and regime]
  {theorem}{strongComp}
\label{thm:strongcomp}
Let $a\in A$ satisfy the precondition of Theorem~\ref{thm:restrict}.
For each $e\in\mathrm{eff}(a)$, suppose $R_e$ admits a strong goal
policy. Let $w_e$ be the optimal worst-case cost of $R_e$, with the
following convention: $w_e\in\mathbb{N}$ if an acyclic strong goal policy exists, and $w_e=\bowtie$ if strong goal policies exist but all are cyclic.

Assume that, for each $e$, there is a strong goal policy $\Pi_e$
attaining $w_e$, and that the family $\{\Pi_e\}_{e\in\mathrm{eff}(a)}$
can be chosen with pairwise-disjoint domains $\sigma(\Pi_e)$. Then
$\Pi=\bigcup_e\Pi_e$ is a strong universal uniform reverse policy for
$a$ in the restricted compilation. Moreover:
\begin{enumerate}
  \item If $w_e\in\mathbb{N}$ for every $e$, then
  $
    k^{\star,\mathsf{u}}_{\mathrm{str,res}}(a)
    =
    \max_{e\in\mathrm{eff}(a)} w_e,
  $
  and therefore
  $
    k^{\star,\mathsf{u}}_{\mathrm{str,full}}(a,S(V))
    \le
    \max_{e\in\mathrm{eff}(a)} w_e.
  $

  \item If $w_e=\bowtie$ for some $e$, then the constructed restricted
  universal uniform policy $\Pi$ is strong cyclic and
  $
    \mathrm{wrc}_{S(V)}(\Pi,a)=\bowtie.
  $
  This certifies the strong-cyclic regime for the restricted compilation,
  but not necessarily for the full problem.
\end{enumerate}
\end{restatable}

The theorem (proof in Appendix~\ref{appx:thmproofs}) is a positive certificate. If the selected per-effect policies
are pairwise disjoint, their union certifies a strong universal uniform
reverse policy. The converse is not claimed. If the selected policies fail
the disjointness condition, this does not prove that no finite strong
universal uniform reverse policy exists in the restricted compilation:
another optimal policy, or a non-optimal policy with larger local cost, may
avoid the conflicting states and merge successfully. Additionally, if some $R_e$ has no strong goal policy, then this compilation cannot
certify strong universal uniform reversibility. This is distinct from the
case $w_e=\bowtie$, where strong policies exist but all restricted strong
policies are cyclic.

\subsection{Certifying non-reversibility within budget $k$}
\label{sub:irrcomp}

The compilations above compute budgets when reversal can be certified. We
now give a complementary sound test for showing that no reversal within a
given budget $k$ exists. This is a lower-bound certificate: if even an
admissible lower bound on returning from some outcome exceeds $k$, then no
actual reverse trajectory for that outcome can have cost at most $k$.

This test is best understood as certifying failure of $k$-bounded weak
reversal. It should not be confused with qualitative $S$-irreversibility:
an action may fail to be reversible within budget $k$ but still be
reversible with a larger budget.

For a state $s\in S(V)$, let
$
  c^\ast_{\mathrm{pre}(a)}(s)
$
denote the minimum cost of any finite trajectory from $s$ to some state
$s'$ such that $\mathrm{pre}(a)\subseteq s'$. If no such trajectory
exists, then $c^\ast_{\mathrm{pre}(a)}(s)=+\infty$.

A heuristic
$
  h:S(V)\to \mathbb{N}\cup\{+\infty\}
$
is admissible for re-achieving $\mathrm{pre}(a)$ if, for every state
$s\in S(V)$,
$
  h(s)\le c^\ast_{\mathrm{pre}(a)}(s)
$.
Thus $h(s)=+\infty$ is admissible only when no finite trajectory from $s$
can reach a state satisfying $\mathrm{pre}(a)$.

\paragraph{Per-state test.}
Let $s$ be relevant, i.e.\ $\mathrm{pre}(a)\subseteq s$, let
$e\in\mathrm{eff}(a)$, and let
$
  s'_e=\gamma(s,a^d_e)
$.
Let $h$ be an admissible heuristic for re-achieving
$\mathrm{pre}(a)$~\cite{helmert2007flexible}. Since returning to the
original state $s$ implies reaching a state satisfying $\mathrm{pre}(a)$,
we have
$
  c^\ast_{\mathrm{pre}(a)}(s'_e)
  \le
  c^\ast(s'_e\to s)
$,
where $c^\ast(s'_e\to s)$ is the minimum cost of any finite trajectory
from $s'_e$ to $s$, and is $+\infty$ if no such trajectory exists.
Therefore,
$$
  h(s'_e)
  \le
  c^\ast_{\mathrm{pre}(a)}(s'_e)
  \le
  c^\ast(s'_e\to s).
$$
Hence, if $h(s'_e)>k$, then
$
  c^\ast(s'_e\to s)>k
$,
and the outcome $e$ cannot be weakly reverted from $s$ within budget
$k$.

\paragraph{Universal test.}
To certify the same conclusion for every relevant state at once, we use a
heuristic that factors through a variable abstraction. Let
$\mathrm{vars}(a)\subseteq V'\subseteq V$, and write
$
  \rho_{V'}:S(V)\to S(V')
$
for the projection of a state onto $V'$.

\begin{definition}[Factored admissible reverse heuristic]
\label{def:factored-admissible-reverse-heuristic}
Let $\mathrm{vars}(a)\subseteq V'\subseteq V$. A heuristic
$
  \hat h:S(V')\to \mathbb{N}\cup\{+\infty\}
$
is admissible for re-achieving $\mathrm{pre}(a)$ through $V'$ if, for
every concrete state $s\in S(V)$,
$
  \hat h(\rho_{V'}(s))
  \le
  c^\ast_{\mathrm{pre}(a)}(s)
$.
Equivalently, the concrete heuristic
$
  h(s)=\hat h(\rho_{V'}(s))
$
is admissible for re-achieving $\mathrm{pre}(a)$.
\end{definition}

In practice, such a heuristic can be obtained from an abstraction whose
abstract transitions simulate all projected concrete transitions with cost
no greater than the corresponding concrete action cost. The optimal
abstract cost-to-go to $\mathrm{pre}(a)|_{V'}$ is then admissible in the
sense of Definition~\ref{def:factored-admissible-reverse-heuristic}.

For $e\in\mathrm{eff}(a)$, define the set of abstract successor states
$$
  \Delta_e^{V'}
  =
  \big\{
    \rho_{V'}(\gamma(s,a^d_e))
    \mid
    s\in S(V),\ \mathrm{pre}(a)\subseteq s
  \big\}.
$$
This set contains exactly the possible projections, onto $V'$, of the
states that can result from applying outcome $e$ of $a$ in any relevant
state. It is a singleton exactly when every variable in $V'$ is fixed by
$\mathrm{pre}(a)$ or by the effect $e$. Otherwise, it ranges over the
free assignments to variables in
$V'\setminus(\mathrm{vars}(\mathrm{pre}(a))\cup\mathrm{vars}(e))$.

\begin{restatable}[Universal failure of $k$-bounded weak reversal]
  {theorem}{irrComp}
\label{thm:irrcomp}
Let $k\in\mathbb{N}$, let $\mathrm{vars}(a)\subseteq V'\subseteq V$, and
let
$
  \hat h:S(V')\to \mathbb{N}\cup\{+\infty\}
$
be admissible for re-achieving $\mathrm{pre}(a)$ through $V'$. If, for
some $e\in\mathrm{eff}(a)$,
$
  \min_{\bar s'\in\Delta_e^{V'}} \hat h(\bar s') > k
$,
then for every state $s\in S(V)$ with $\mathrm{pre}(a)\subseteq s$, the
outcome
$
  \gamma(s,a^d_e)
$
cannot be weakly reverted from $s$ within budget $k$. Hence $a$ is not
$k$-bounded weakly universally reversible.

If the minimum is $+\infty$, i.e.\
$
  \hat h(\bar s')=+\infty
  \quad\text{for every } \bar s'\in\Delta_e^{V'}
$,
then this outcome cannot be weakly reverted from any relevant state at any
finite budget.
\end{restatable}

Theorem~\ref{thm:irrcomp} (proof in Appendix~\ref{appx:thmproofs}) is sound but incomplete. If the abstract lower
bound exceeds $k$, then $k$-bounded weak universal reversibility is
impossible. If the abstract lower bound is at most $k$, no conclusion
follows: the action may or may not be reversible within budget $k$. The
test is useful as a fast reject before running the cost-optimal searches
of Sections~\ref{sub:weakcomp} and~\ref{sub:strongcomp}.

\subsection{The advantage of a cost-optimal compilation}
\label{sub:boand}




Cost-optimal classical planning yields an exact restricted weak budget,
while worst-case-optimal FOND search distinguishes finite strong budgets
from strong-cyclic witnesses with unbounded worst-case cost. The
admissible lower-bound test can additionally reject reversal within a
proposed budget $k$. Together these procedures provide quantitative
certificates beyond qualitative reversibility, although an inconclusive
test does not settle the full-domain problem.

\section{Experiments}
\label{sec:experiments}

We evaluate each grounded action in the FOND benchmark suite of
\citet{chrpa2024weak} as a candidate for reversal. The suite contains
$1871$ actions, of which $1178$ satisfy the structural condition required
by our restricted compilations. We use three positive action-cost models:
unit costs $c_{\mathrm{unit}}(a)=1$, effect-size costs
$c_{\mathrm{eff}}(a)=1+|\mathrm{vars}(a)|$, and outcome-sensitive costs
$c_{\mathrm{nd}}(a)=1+|\mathrm{vars}(a)|+|\mathrm{eff}(a)|-1$.
Implementation details, the full protocol, and additional results appear
in Appendix~\ref{app:experiments}.

\paragraph{Classification.}
Table~\ref{tab:classification} reports the action classifications under
unit costs; the classifications are unchanged under the other two cost
models. The weak, finite-strong, and strong-cyclic columns report
certificates from the restricted compilation, not a complete
classification of the full-domain problem.

\begin{table}[t]
\centering
\small
\setlength{\tabcolsep}{2.5pt}
\begin{tabular}{lrrrrrrrrr}
\toprule
Domain
& $|A|$
& $\subseteq$
& $\mathsf{Wk}$
& $\mathsf{BStr}$
& $\mathsf{SCyc}$
& $\neg\mathsf{Str}$
& \textsc{MF}
& \textsc{I}
& \textsc{?}\\
\midrule
\begin{filecontents*}{results/table1_unit.csv}
Domain,nA,nSubset,nNotSubset,nWkRes,nBStrRes,nSCycRes,nNoStrong,nMergeFail,nTimeout,nIRR,nUnknown
acrobatics,5,5,0,4,3,0,0,1,0,0,1
beam-walk,7,7,0,6,0,0,6,0,0,1,0
blocksworld,190,5,185,5,5,0,0,0,0,0,185
bus-fare,5,5,0,2,0,2,0,0,0,3,0
climber,3,3,0,0,0,0,0,0,0,3,0
doors,5,0,5,0,0,0,0,0,0,4,1
earth-observation,27,27,0,21,21,0,0,0,0,6,0
elevators,41,8,33,8,8,0,0,0,0,3,30
ex-blocksworld,85,5,80,0,0,0,0,0,0,35,50
faults,51,0,51,0,0,0,0,0,0,25,26
first-responders,52,42,10,28,28,0,0,0,0,12,12
forest,156,30,126,10,6,0,4,0,0,16,130
islands,24,24,0,20,20,0,0,0,0,4,0
miner,179,176,3,158,158,0,0,0,0,18,3
river,3,1,2,0,0,0,0,0,0,3,0
spiky-tireworld,211,130,81,124,124,0,0,0,0,3,84
tireworld,52,44,8,0,0,0,0,0,0,7,45
tireworld-truck,24,14,10,10,10,0,0,0,0,4,10
triangle-tireworld,11,8,3,0,0,0,0,0,0,11,0
zenotravel,740,644,96,504,0,504,0,0,0,0,236
Total,1871,1178,693,900,383,506,10,1,0,158,813
\end{filecontents*}
\csvreader[
  head to column names,
  late after line=\\
]{results/table1_unit.csv}{
  1=\csvDomain,
  2=\csvA,
  3=\csvSubset,
  4=\csvNotSubset,
  5=\csvWkRes,
  6=\csvBStrRes,
  7=\csvSCycRes,
  8=\csvNoStrong,
  9=\csvMergeFail,
  10=\csvTimeout,
  11=\csvIRR,
  12=\csvUnknown
}{%
{\scriptsize\textsc{%
\ifcsvstrcmp{\csvDomain}{earth-observation}{earth-obs.}{%
\ifcsvstrcmp{\csvDomain}{first-responders}{first-resp.}{%
\ifcsvstrcmp{\csvDomain}{ex-blocksworld}{ex.-blocks.}{%
\ifcsvstrcmp{\csvDomain}{spiky-tireworld}{spiky-tire.}{%
\ifcsvstrcmp{\csvDomain}{tireworld-truck}{tire.-truck}{%
\ifcsvstrcmp{\csvDomain}{triangle-tireworld}{tri.-tire.}{%
\ifcsvstrcmp{\csvDomain}{blocksworld}{blocks.}{%
\ifcsvstrcmp{\csvDomain}{zenotravel}{zeno.}{%
\csvDomain
}}}}}}}}%
}}
& \csvA
& \csvSubset
& \csvWkRes
& \csvBStrRes
& \csvSCycRes
& \csvNoStrong
& \csvMergeFail
& \csvIRR
& \csvUnknown
}%
\bottomrule
\end{tabular}
\caption{Classification of actions under unit costs. Domain names are
abbreviated; full names appear in Appendix~\ref{app:experiments}.
The weak, finite-strong, and strong-cyclic columns report certificates
in the restricted compilation.}
\label{tab:classification}
\vspace{-0.2in}
\end{table}

Overall, $900$ actions receive a restricted weak universal uniform
certificate. Of these, $383$ have finite restricted strong worst-case
budgets, $506$ have constructed restricted strong-cyclic witnesses, $10$
have no restricted strong witness, and $1$ results in \textsc{MergeFail}.
Another $158$ actions are certified universally irreversible; $813$
remain unclassified by the sufficient tests. No planner call times out.
The strong-cyclic witnesses guarantee recovery under fairness but have
unbounded worst-case finite recovery cost.

\paragraph{Reversal budgets.}
Table~\ref{tab:budgets} reports finite restricted budgets across the
three cost models. Weak budgets measure cheapest successful recovery
from each outcome, whereas strong budgets bound the worst terminal
trajectory. Strong-cyclic witnesses with value $\bowtie$ are excluded
from the finite-budget statistics.

\begin{table}[t]
\centering
\small
\setlength{\tabcolsep}{2pt}
\begin{tabular*}{\columnwidth}{@{\extracolsep{\fill}}llrrrrrrrr@{}}
\toprule
Cost
& Type
& $\#$
& Min
& Q1
& Med.
& Mean
& Q3
& Max
& Std.\\
\midrule
\begin{filecontents*}{results/table2.csv}
CostModel,Budget,N,Min,Qone,Median,Mean,Qthree,Max,Std
unit,wk,900,0,1,5.0,4.499,7,9,3.336
unit,str,383,0,1,1,1.023,1,2,0.293
effect,wk,900,0,4,17.0,17.246,28,35,13.42
effect,str,383,0,3,4,3.428,4,4,0.83
outcome,wk,900,0,4,20.0,19.561,32,40,15.526
outcome,str,383,0,3,4,3.428,4,4,0.83
\end{filecontents*}
\csvreader[
  head to column names,
  late after line=\\
]{results/table2.csv}{}{%
\ifcsvstrcmp{\CostModel}{unit}{Unit}{%
  \ifcsvstrcmp{\CostModel}{effect}{Eff.}{Out.}%
}
&
\ifcsvstrcmp{\Budget}{wk}{Wk}{Str}
& \N
& \Min
& \Qone
& \Median
& \Mean
& \Qthree
& \Max
& \Std
}%
\bottomrule
\end{tabular*}
\caption{Distribution of finite restricted reversal budgets across
certified actions. Wk and Str denote
$k^{\star,\mathsf{u}}_{\mathrm{wk,res}}$ and
$k^{\star,\mathsf{u}}_{\mathrm{str,res}}$, respectively;
Eff. and Out. denote effect-size and outcome-sensitive costs.}
\label{tab:budgets}
\vspace{-0.2in}
\end{table}

Under unit costs, the median finite weak budget is $5$ and the median
finite strong budget is $1$. This does not mean that strong reversal is
generally cheaper: cyclic strong witnesses are excluded from the finite
strong-budget distribution. The mean finite weak budget rises from
$4.499$ under unit costs to $17.246$ under effect-size costs and
$19.561$ under outcome-sensitive costs. Thus the cost model changes the
recovery budget even when the qualitative classification is unchanged.

\paragraph{Scope of the results.}
Finite budgets are exact for the restricted compilations but only upper
bounds on the corresponding full-domain optimal budgets. Likewise, a
constructed restricted strong-cyclic witness does not establish that every
full-domain strong witness is cyclic. Appendix~\ref{app:experiments}
provides per-domain budgets, cost-sensitivity ratios, lower-bound rejection
results, runtimes, and further limitations.

\section{Conclusion}

Cost-sensitive reversibility distinguishes actions that can be
recovered cheaply (possibly or surely), actions requiring a larger recovery budget, and
actions whose strong-cyclic recovery has unbounded worst-case cost.
We established the corresponding budget notions and merging results,
and evaluated restricted compilations and a lower-bound rejection
test on FOND benchmarks. The computed finite budgets are exact in
the restricted search space but only upper bounds on full-domain
optima; restricted strong cyclicity likewise does not rule out a
finite full-domain witness. A next step is to search beyond the
restricted space, including directly for mergeable families of
per-effect policies.
\section*{Disclaimer}
This paper was prepared for informational purposes by the Artificial Intelligence Research group of JPMorgan Chase \& Co. and its affiliates ("JP Morgan'') and is not a product of the Research Department of JP Morgan. JP Morgan makes no representation and warranty whatsoever and disclaims all liability, for the completeness, accuracy or reliability of the information contained herein. This document is not intended as investment research or investment advice, or a recommendation, offer or solicitation for the purchase or sale of any security, financial instrument, financial product or service, or to be used in any way for evaluating the merits of participating in any transaction, and shall not constitute a solicitation under any jurisdiction or to any person, if such solicitation under such jurisdiction or to such person would be unlawful.

© 2026 JPMorgan Chase \& Co. All rights reserved

\bibliography{bibliography}


\appendix
\onecolumn
\section{Proofs}\label{appx:proofs}
\subsection{Propositions}\label{appx:propproofs}
\acyclicFinite*
\begin{proof}
Suppose first that $\mathcal{P}$ is bounded. Then, for every relevant
$s\in S_a$ and every outcome $s'\in\delta(s,a)$, the graph reachable from
$s'$ under $\Pi_s$ is finite and acyclic. Hence every terminal trajectory
from $s'$ is a simple path of bounded length, and because action costs are
positive and finite, the maximum terminal trajectory cost is finite. Taking
the maximum over the finitely many relevant pairs $(s,s')$ yields
$\mathrm{wrc}_S(\mathcal{P},a)\in\mathbb{N}$.

Conversely, if $\mathcal{P}$ is not bounded, then for some relevant
$s\in S_a$ and some $s'\in\delta(s,a)$, a cycle is reachable from $s'$
under $\Pi_s$. Since all action costs are strictly positive, the cycle can
be traversed arbitrarily many times before returning to the target under
the strong-cyclic interpretation, yielding terminal trajectories of
unbounded finite cost. Thus
$\mathrm{worst}_s(\Pi_s,s')=\bowtie$, and consequently
$\mathrm{wrc}_S(\mathcal{P},a)=\bowtie$.
\end{proof}

\refinedLattice*
\begin{proof}
(1) A witness satisfying a budget $k$ also satisfies any larger budget $k'$.

(2) Let $\mathcal{P}=\{\Pi_s\}_{s\in S_a}$ witness
$\mathsf{Str}^{\mu}_k(a,S)$. Since the witness is admissible in mode
$\mu$ and strong reversal implies weak reversal state by state,
$\mathcal{P}$ is also a weak witness admissible in mode $\mu$. Moreover,
for every relevant $s\in S_a$ and every $s'\in\delta(s,a)$,
$$
  \mathrm{best}_s(\Pi_s,s')\le \mathrm{worst}_s(\Pi_s,s').
$$
Taking maxima over all relevant pairs gives
$$
  \mathrm{brc}_S(\mathcal{P},a)
  \le
  \mathrm{wrc}_S(\mathcal{P},a)
  \le k.
$$
Hence $\mathsf{Wk}^{\mu}_k(a,S)$ holds.

(3) This follows directly from Definition~\ref{def:bounded-predicates}.
If $a$ is strongly reversible in mode $\mu$, then either some admissible
strong witness is bounded, in which case $\mathsf{BStr}^{\mu}(a,S)$ holds,
or no admissible bounded strong witness exists, in which case
$\mathsf{SCyc}^{\mu}(a,S)$ holds. The two cases are mutually exclusive by
definition.

(4) The implication
$\mathsf{BWk}^{\mu}(a,S)\Rightarrow\mathsf{WkRev}^{\mu}(a,S)$ is immediate.
For the converse, suppose $\mathsf{WkRev}^{\mu}(a,S)$ holds, witnessed by
an admissible weak witness $\mathcal{P}=\{\Pi_s\}_{s\in S_a}$. For every
relevant $s\in S_a$ and every outcome $s'\in\delta(s,a)$, weak reversal
gives a finite terminal trajectory from $s'$ back to $s$ under $\Pi_s$.
Since action costs are positive and the trajectory is finite,
$\mathrm{best}_s(\Pi_s,s')\in\mathbb{N}$. There are finitely many relevant
pairs $(s,s')$, so $\mathrm{brc}_S(\mathcal{P},a)\in\mathbb{N}$. Therefore
$\mathsf{Wk}^{\mu}_k(a,S)$ holds for
$k=\mathrm{brc}_S(\mathcal{P},a)$.

(5) Bounded strong reversibility implies strong reversibility by
definition. Strong reversibility implies weak reversibility by item~(2),
ignoring the budget. Finally, weak reversibility in either mode implies
$\neg\,\mathsf{Irr}(a,S)$, because a weak witness provides, for every
relevant state $s\in S_a$, at least one policy that weakly reverts $a$ from
$s$. This contradicts the definition of $S$-irreversibility, which requires
that no relevant state admit such a policy.

The converse of the last implication fails in general. If $|S_a|\ge 2$, it
may happen that some relevant states admit weak reverting policies while
others do not. Then $a$ is not $S$-irreversible, but it is also not weakly
$S$-reversible, because weak $S$-reversibility requires a witness for every
relevant state.
\end{proof}

\subsection{Theorems}\label{appx:thmproofs}

\weakMergingCosts*
\begin{proof}
Since policies are relations, $\Pi_1\cup\Pi_2$ is again a relation over
states and applicable actions, hence a policy.

Fix $s\in S_1^a$ and $s'\in\delta(s,a)$. Since $\Pi_1$ weakly reverts
$a$ from $s$, there exists a finite $\Pi_1$-trajectory from $s'$ to
the terminal state $s$. Every transition of this trajectory is also a
transition of $\Pi_1\cup\Pi_2$. Moreover, by \textup{(C1)}, the target
state $s$ is not in $\sigma(\Pi_2)$; since $s$ is terminal under
$\Pi_1$, it is also terminal under $\Pi_1\cup\Pi_2$. Hence
$
  s\in\tau(\Pi_1\cup\Pi_2,s').
$
The same argument applies symmetrically to every $s\in S_2^a$. Therefore
$\Pi_1\cup\Pi_2$ is a weak uniform $(S_1\cup S_2)$-reverse policy for
$a$.

For the cost bound, fix again $s\in S_1^a$ and $s'\in\delta(s,a)$. The
successful terminal trajectories from $s'$ to $s$ under $\Pi_1$ are
also successful terminal trajectories from $s'$ to $s$ under
$\Pi_1\cup\Pi_2$. Therefore the cheapest successful reversal to $s$
cannot become more expensive:
$$
  \mathrm{best}_{s}(\Pi_1\cup\Pi_2,s')
  \le
  \mathrm{best}_{s}(\Pi_1,s')
  \le k_1.
$$
Symmetrically, for every $s\in S_2^a$ and $s'\in\delta(s,a)$,
$$
  \mathrm{best}_{s}(\Pi_1\cup\Pi_2,s')
  \le k_2.
$$
Taking the maximum over all relevant $s\in S_1^a\cup S_2^a$ and
$s'\in\delta(s,a)$ gives
$$
  \mathrm{brc}_{S_1\cup S_2}(\Pi_1\cup\Pi_2,a)
  \le
  \max(k_1,k_2).
$$
\end{proof}

\begin{restatable}[Iterated weak merging]
  {corollary}{iteratedWeakMerging}
\label{cor:weak_merging}
Let $\{\Pi_i\}_{i=1}^{n}$ be weak uniform $S_i$-reverse policies for
$a$, where $\Pi_i$ is $k_i$-bounded. Suppose that the policies
pairwise satisfy the weak non-interference condition, i.e.,
$
  \sigma(\Pi_j)\cap S_i^a=\emptyset
  \qquad\text{for all } i\neq j
$.
Then
$
  \Pi=\bigcup_{i=1}^{n}\Pi_i
$
is a weak uniform $\big(\bigcup_i S_i\big)$-reverse policy for $a$, and
$
  \mathrm{brc}_{\bigcup_i S_i}(\Pi,a)
  \le
  \max_i k_i
$.
\end{restatable}
\begin{proof}
By induction on $n$, using Theorem~\ref{thm:weak_merging}. The pairwise
non-interference assumptions ensure that, at each induction step, the
current union and the next component satisfy \textup{(C1)}.
\end{proof}

\begin{remark}[Execution of relational weak policies]
The relational policy $\Pi_1\cup\Pi_2$ certifies the existence of a
bounded weak reversal. If one wants to execute a deterministic controller,
one may need to resolve the action choices in states related to multiple
actions. Such a resolution can be chosen with the current reversal target
$s$ as context, selecting an action that lies on a successful
target-reaching branch. This execution issue is separate from the
existence and budget certificate proved above.
\end{remark}

\strongMergingCosts*
\begin{proof}
First, $\Pi_1\cup\Pi_2$ is a policy because policies are relations and
the union of relations is a relation over states and applicable actions.

We prove confinement. Fix $s\in S_1^a$ and $s'\in\delta(s,a)$. We show
that every trajectory of $\Pi_1\cup\Pi_2$ starting from $s'$ uses only
actions of $\Pi_1$ until it terminates at $s$. Since $\Pi_1$ strongly
reverts $a$ from $s$, every state reachable from $s'$ under $\Pi_1$
is either the unique terminal $s$, or is a non-terminal state in
$\sigma(\Pi_1)$. The target $s$ is not in $\sigma(\Pi_2)$ by
\textup{(C1)}, and every state in $\sigma(\Pi_1)$ is outside
$\sigma(\Pi_2)$ by \textup{(C2)}. Hence no action of $\Pi_2$ is
available, via the merged policy, along a reversal owned by $\Pi_1$.
Thus the reachable transition structure from $s'$ under
$\Pi_1\cup\Pi_2$ coincides with that under $\Pi_1$.

The same argument applies symmetrically to every $s\in S_2^a$ and
$s'\in\delta(s,a)$. Therefore, for each relevant state in
$S_1\cup S_2$, the merged policy has the same reachable terminals and
same reachability-to-target property as the component that owns that state.
Hence $\Pi_1\cup\Pi_2$ is a strong uniform $(S_1\cup S_2)$-reverse
policy for $a$.

The cost claims follow from the same confinement property. For
$s\in S_i^a$ and $s'\in\delta(s,a)$,
$$
  \mathrm{worst}_{s}(\Pi_1\cup\Pi_2,s')
  =
  \mathrm{worst}_{s}(\Pi_i,s').
$$
Therefore, if both components are bounded, the global worst-case reversal
cost of the merge is exactly the maximum of the two component worst-case
reversal costs:
$$
  \mathrm{wrc}_{S_1\cup S_2}(\Pi_1\cup\Pi_2,a)
  =
  \max\big(
    \mathrm{wrc}_{S_1}(\Pi_1,a),
    \mathrm{wrc}_{S_2}(\Pi_2,a)
  \big).
$$
If at least one component is strong cyclic, then for some component
$i$, some relevant $s\in S_i^a$, and some $s'\in\delta(s,a)$, we
have
$$
  \mathrm{worst}_{s}(\Pi_i,s')=\bowtie.
$$
By confinement, the same unbounded reversal is present in the merge:
$$
  \mathrm{worst}_{s}(\Pi_1\cup\Pi_2,s')=\bowtie.
$$
Hence the merged policy is strong cyclic and its global worst-case reversal
cost is $\bowtie$.
\end{proof}

\begin{remark}[No additional no-cross-cycle condition]
Conditions \textup{(C1)} and \textup{(C2)} already imply that a reversal
owned by one component cannot enter states controlled by the other
component. Thus the merge cannot create new cross-component cycles along
relevant strong reversals. No separate no-cross-cycle hypothesis is needed.
\end{remark}

\begin{restatable}[Iterated strong merging]
  {corollary}{iteratedStrongMerging}
\label{cor:strong_merging}
Let $\{\Pi_i\}_{i=1}^{n}$ be strong uniform $S_i$-reverse policies for
$a$. Suppose they pairwise satisfy \textup{(C1)} and \textup{(C2)}. Then
$
  \Pi=\bigcup_{i=1}^{n}\Pi_i
$
is a strong uniform $\big(\bigcup_i S_i\big)$-reverse policy for $a$.
Moreover,
\begin{align*}
  \mathrm{wrc}_{\bigcup_i S_i}&(\Pi,a)
  =\begin{cases}
  \max_i \mathrm{wrc}_{S_i}(\Pi_i,a),
    & \text{if all } \Pi_i \text{ are bounded},\\[1mm]
  \bowtie,
    & \text{if some } \Pi_i \text{ is strong cyclic}.
  \end{cases}
\end{align*}
In particular, if all $\Pi_i$ are $k_i$-bounded, then
$
  \mathrm{wrc}_{\bigcup_i S_i}(\Pi,a)
  \le
  \max_i k_i
$.
\end{restatable}

\begin{proof}
By induction on $n$, using Theorem~\ref{thm:strong_merging}. Pairwise
\textup{(C1)} ensures that no component assigns actions to another
component's relevant targets, and pairwise \textup{(C2)} ensures that the
domains of all component policies are disjoint. Hence the confinement
argument applies at each induction step.
\end{proof}

\begin{remark}[Strong/weak asymmetry]
Weak merging and strong merging differ structurally. In the weak case,
adding relational branches preserves successful target-reaching
trajectories and may introduce cheaper ones, so the best-case budget
composes as an upper bound:
$
  \mathrm{brc}_{S_1\cup S_2}(\Pi_1\cup\Pi_2,a)
  \le
  \max(k_1,k_2)
$.
In the strong case, additional branches may invalidate guarantees unless
they are excluded from the relevant reachable regions. Conditions
\textup{(C1)} and \textup{(C2)} provide this confinement. Consequently,
the worst-case cost of the merge is inherited exactly from the component
owning each relevant target, and the global regime is bounded iff all
components are bounded.
\end{remark}

\costInvariantPrecheck*
\begin{proof}
By Lemma~9 of~\citet{chrpa2024weak}, the hypothesis implies that no strong
universal uniform reverse policy exists. A finite-budget strong universal
uniform reverse policy would in particular be a strong universal uniform
reverse policy. Likewise, a strong-cyclic universal uniform witness would
also presuppose such a qualitative strong witness. Hence neither kind of
strong universal uniform witness exists.
\end{proof}

\costRestrict*
\begin{proof}
Universality is exactly the content of Theorem~10 of
\citet{chrpa2024weak}: since the action $a$, and all actions used by the
reverse policy, affect only variables in $\mathrm{vars}(\mathrm{pre}(a))$,
all variables outside this set remain invariant and cannot affect the
applicability or outcome of the restricted reverse policy. Therefore a
restricted witness works from every full state satisfying
$\mathrm{pre}(a)$.

For the budget claim, the restricted policy is also an admissible full-domain
universal uniform witness of the same cost. Since the full optimum ranges
over all full-domain witnesses, including this one, the full optimum cannot
be larger than the restricted witness cost.
\end{proof}

\weakComp*
\begin{proof}
For each determinization $a^d_e$, the task $T^d_e$ asks for a cheapest
plan returning from $\mathrm{ha}(a^d_e)$ to $\mathrm{pre}(a)$ in the
restricted variable space. Since every $T^d_e$ is solvable, Theorem~11 of
\citet{chrpa2024weak} yields, for each $e$, a policy $\Pi_e$ induced by
$\pi_e$ that weakly reverts the corresponding outcome. Each $\Pi_e$
uses only variables and actions in the restricted domain.

The policies $\Pi_e$ all return to the same restricted target
$\mathrm{pre}(a)$. Moreover, none assigns an action to that target, so the
weak merging theorem applies iteratively. Thus
$\Pi=\bigcup_e\Pi_e$ is a weak universal uniform reverse policy.

For the cost, fix a full state $s$ satisfying $\mathrm{pre}(a)$ and an
effect $e\in\mathrm{eff}(a)$, and let $
  s'=\gamma(s,a^d_e).
$
In the restricted task, the start is $\mathrm{ha}(a^d_e)$ and the target is
$\mathrm{pre}(a)$. Since both $a^d_e$ and all actions in the restricted
policy affect only variables in $\mathrm{vars}(\mathrm{pre}(a))$, every full
outcome $s'$ produced by effect $e$ projects to the same restricted start
$\mathrm{ha}(a^d_e)$, while variables outside
$\mathrm{vars}(\mathrm{pre}(a))$ remain fixed at their values in $s$. Thus
a restricted trajectory from $\mathrm{ha}(a^d_e)$ to $\mathrm{pre}(a)$ lifts
to a full trajectory from $s'$ back to $s$. Moreover, any cheaper
target-reaching trajectory introduced by the merge would project to a
cheaper plan for $T^d_e$, contradicting the optimality of $\pi_e$. Hence
$$
  \mathrm{best}_{s}(\Pi,s')=c(\pi_e).
$$
Therefore the value of $\mathrm{best}_{s}(\Pi,s')$ depends only on the
effect $e$ producing $s'$, and not on the particular full state $s$; hence
maximizing over all relevant states $s$ and outcomes $s'\in\delta(s,a)$ is
equivalent to maximizing over effects $e\in\mathrm{eff}(a)$. Thus
$$
  \mathrm{brc}_{S(V)}(\Pi,a)
  =
  \max_{e\in\mathrm{eff}(a)} c(\pi_e).
$$
Optimality of each $\pi_e$ gives exactness for the restricted compilation.
The full-domain upper bound then follows from Theorem~\ref{thm:restrict}.
\end{proof}

\strongComp*
\begin{proof}
Each task $R_e$ ranges only over $\mathrm{vars}(\mathrm{pre}(a))$, and
each strong goal policy $\Pi_e$ strongly reverts the corresponding
determinization $a^d_e$ in the restricted space. Since the policies have pairwise-disjoint domains, the strong merging theorem applies.
Condition \textup{(C2)} is the assumed disjointness. Condition \textup{(C1)}
holds because all policies return to the same restricted target
$\mathrm{pre}(a)$, and a strong reverse policy does not assign an action
to its terminal state. Hence $\Pi=\bigcup_e\Pi_e$ is a strong universal
uniform reverse policy in the restricted compilation.

If all $w_e$ are finite, then every $\Pi_e$ is acyclic. By the strong
merging theorem,
$$
  \mathrm{wrc}_{S(V)}(\Pi,a)
  =
  \max_{e\in\mathrm{eff}(a)} w_e.
$$
Optimality of each $w_e$ gives exactness for the restricted compilation.
The full-domain upper bound follows from Theorem~\ref{thm:restrict}.

If $w_e=\bowtie$ for some $e$, then every strong policy for $R_e$ is
cyclic. The selected $\Pi_e$ is therefore strong cyclic. By the strong
merging theorem, the merged policy $\Pi$ is strong cyclic and has
$$
  \mathrm{wrc}_{S(V)}(\Pi,a)=\bowtie.
$$
This conclusion concerns the restricted compilation. An unrestricted
full-domain policy might still avoid the cycle by using actions outside
$A_{\mathrm{pre}}$, so the result does not by itself imply
$\mathsf{SCyc}^{\mathsf{u}}(a,S(V))$ in the full domain.
\end{proof}

\irrComp*
\begin{proof}
Fix $e\in\mathrm{eff}(a)$ satisfying the hypothesis, and let $s$ be any
state with $\mathrm{pre}(a)\subseteq s$. Put
$
  s'_e=\gamma(s,a^d_e)
$.
By definition of $\Delta_e^{V'}$,
$
  \rho_{V'}(s'_e)\in \Delta_e^{V'}
$.
Therefore, by the hypothesis,
$
  \hat h(\rho_{V'}(s'_e))
  \ge
  \min_{\bar s'\in\Delta_e^{V'}}\hat h(\bar s')
  >
  k
$.
Since $\hat h$ is admissible for re-achieving $\mathrm{pre}(a)$ through
$V'$,
$
  \hat h(\rho_{V'}(s'_e))
  \le
  c^\ast_{\mathrm{pre}(a)}(s'_e)
$.
Thus
$
  c^\ast_{\mathrm{pre}(a)}(s'_e)>k
$.

Now consider any finite trajectory from $s'_e$ back to $s$. Since
$\mathrm{pre}(a)\subseteq s$, such a trajectory is also a trajectory from
$s'_e$ to a state satisfying $\mathrm{pre}(a)$. Hence its cost is at least
$c^\ast_{\mathrm{pre}(a)}(s'_e)$, and therefore greater than $k$. Thus no
trajectory from $s'_e$ to $s$ has cost at most $k$.

Since $s$ was arbitrary, the outcome $e$ cannot be weakly reverted within
budget $k$ from any relevant state. Consequently, $a$ is not
$k$-bounded weakly universally reversible.

If the minimum is $+\infty$, then the same argument applies for every
finite $k$, so no finite budget suffices for that outcome from any
relevant state.
\end{proof}
\section{Additional Experimental Details}
\label{app:experiments}

\subsection{Research questions and benchmarks}

The experiments address five questions: (RQ1) how often the restricted
compilations certify weak or finite strong universal uniform reversal;
(RQ2) how often a certified weak action also has a finite strong witness
rather than a restricted strong-cyclic witness; (RQ3) how budgets vary
across action-cost models; (RQ4) how effectively admissible lower bounds
reject a proposed budget; and (RQ5) the computational overhead of the
pipeline.

We use the grounded FOND benchmark suite of~\citet{chrpa2024weak},
comprising the following domains:
$$
\begin{array}{llll}
\textsc{acrobatics},&
\textsc{beam-walk},\\
\textsc{blocksworld},&
\textsc{bus-fare},\\
\textsc{climber},&
\textsc{doors},\\
\textsc{earth-observation},&
\textsc{elevators},\\
\textsc{exploding-blocksworld},&
\textsc{faults},\\
\textsc{first-responders},&
\textsc{forest},\\
\textsc{islands},&
\textsc{miner},\\
\textsc{river},&
\textsc{spiky-tireworld},\\
\textsc{tireworld},&
\textsc{tireworld-truck},\\
\textsc{triangle-tireworld},&
\textsc{zenotravel}.
\end{array}
$$
Each lifted domain is instantiated as a grounded FOND representation,
and each grounded action is analysed independently. Across the suite,
$1871$ actions are analysed. Of these, $1178$ satisfy
$
  \bigcup_{e\in\mathrm{eff}(a)}\mathrm{vars}(e)
  \subseteq
  \mathrm{vars}(\mathrm{pre}(a)),
$
and $693$ fail this structural condition. By
Corollary~\ref{cor:precheck}, actions failing it cannot be strongly
universally uniformly reversible. We do not apply the restricted weak
or strong compilations to those actions.

\subsection{Cost models and experimental pipeline}

Most benchmark instances do not provide native action costs. We use
three reproducible positive-cost models:
$$
\begin{aligned}
&c_{\mathrm{unit}}(a) = 1,\\
&c_{\mathrm{eff}}(a) = 1+|\mathrm{vars}(a)|,\\
&c_{\mathrm{nd}}(a) =
  1+|\mathrm{vars}(a)|+|\mathrm{eff}(a)|-1.
\end{aligned}
$$
Unit costs measure the number of recovery actions; effect-size costs
penalize actions with larger syntactic footprints; outcome-sensitive
costs additionally penalize actions with more possible effects. All
three map actions to $\mathbb{N}^{+}$.

For each eligible action $a$, the pipeline has the following steps.

\paragraph{Weak restricted budget.}
For each $e\in\mathrm{eff}(a)$, we construct the cost-optimal classical
task $T^d_e$ of Theorem~\ref{thm:weakcomp}. If every such task is
solvable, the restricted weak universal uniform budget is
$$
  k^{\star,\mathsf{u}}_{\mathrm{wk,res}}(a)
  =
  \max_{e\in\mathrm{eff}(a)}c(\pi_e),
$$
where $\pi_e$ is an optimal plan for $T^d_e$. The value is exact for
the restricted compilation and, by Theorem~\ref{thm:restrict}, is an
upper bound on the full-domain optimal budget.

\paragraph{Strong restricted budget and regime.}
For each effect $e$, we construct the restricted FOND task $R_e$ of
Theorem~\ref{thm:strongcomp} and solve it with a worst-case-optimal
FOND solver. Each task is classified as \textsc{NoStrong} if no strong
goal policy exists, \textsc{Finite} if an acyclic strong policy exists
with optimal worst-case cost $w_e\in\mathbb{N}$, \textsc{Cyclic} if
strong policies exist but all are cyclic ($w_e=\bowtie$), or
\textsc{Timeout} if no conclusive result is obtained within the time
limit.

If every $R_e$ is solved and the selected policies have pairwise-disjoint
domains, we merge them using the iterated strong-merging result following
Theorem~\ref{thm:strong_merging}. When every $w_e$ is finite, the
certified restricted budget is
$$
  k^{\star,\mathsf{u}}_{\mathrm{str,res}}(a)
  =
  \max_{e\in\mathrm{eff}(a)}w_e.
$$
If a selected value is $\bowtie$, the constructed restricted witness
is strong cyclic. This does not establish full-domain strong cyclicity.
If strong per-effect policies are found but the selected family is not
pairwise disjoint, we report \textsc{MF} (merge fail). That result is
inconclusive: another selection of policies might merge.

\paragraph{Lower-bound rejection.}
For selected budgets $k$, we apply Theorem~\ref{thm:irrcomp} using
abstractions $V'$ with
$\mathrm{vars}(a)\subseteq V'\subseteq V$ and admissible heuristics
factoring through $V'$. A rejection certifies that some outcome cannot
be weakly reverted from any relevant state within budget $k$. Failure
to reject is inconclusive.

\paragraph{Universal irreversibility}. We certify universal irreversibility using the projected-planning unsolvability test of~\citet{chrpa2024weak}. This classification is separate from Theorem~\ref{thm:irrcomp}, which rejects reversal only within a specified budget $k$.

\subsection{Implementation and reported metrics}\label{appx:exp_details}

The compilations are implemented as a preprocessing layer taking a
grounded FOND domain and a candidate action as input. Weak tasks are
solved with \textsc{FastDownward}~\cite{helmert2006fast}, using
\textsc{A*} guided by \textsc{LM-Cut}. Strong tasks are solved with our cost-aware extension of \textsc{BOAND*}~\cite{aineto2025cost}, configured for the worst-case metric as first objective and the best-case metric as second. Our extension replaces unit-cost ancestor counting with weighted
graph evaluation after each policy expansion. Action costs are
extracted from PDDL and attached to the grounded determinizations. For the reachable graph induced by a partial policy,
Dijkstra's algorithm computes minimum accumulated costs from the
initial state. Maximum accumulated costs are computed by memoized
recursion over predecessors outside the cyclic region; states
reachable from a detected cycle receive the unbounded-cost value.
These values enter the best- and worst-case frontier bounds used
to order policy expansions and prune dominated candidates.
Completed policies are additionally checked for properness before
being returned. BOAND* returns a policy assigning at most one action to each state. We merge per-effect policies only when their domains are pairwise disjoint, so every strong reverse policy constructed by our pipeline also assigns at most one action per state. 
\paragraph{Preservation of the search bounds.}
Our weighted graph evaluation implements the \textsc{MinSum} and
\textsc{MaxSum} bounds of~\citet{aineto2025cost}.
Let $O(\Pi)$ contain the reachable goal and pending states of a
partial policy $\Pi$, and let $g_{\min}(s)$ and $g_{\max}(s)$
denote the minimum and maximum accumulated costs from the initial
state to $s$. For an admissible all-outcome heuristic $h$, the bounds are
$$
f_{\mathrm{best}}(\Pi)
 = \min_{s\in O(\Pi)}\bigl(g_{\min}(s)+h(s)\bigr),
\qquad
f_{\mathrm{worst}}(\Pi)
 = \max_{s\in O(\Pi)}\bigl(g_{\max}(s)+h(s)\bigr).
$$
Dijkstra's algorithm computes $g_{\min}$ exactly, while
predecessor recursion computes $g_{\max}$ in the acyclic region.
A reachable cycle persists under policy extension because assigned
actions are not replaced; consequently, any proper completion
containing that cycle has unbounded worst-case cost.
Every completed trajectory either reaches an existing goal or
extends a prefix ending at a pending state. Admissibility of $h$
therefore makes the two expressions lower bounds on the respective
costs of every solution extending $\Pi$. On completed proper
policies, all outgoing states are goals and $h$ is zero, so the
bounds equal the corresponding policy costs. Thus weighted
evaluation preserves the admissibility and goal-awareness required
by BOAND*'s optimality argument
\cite[Lemma~2 and Theorem~2]{aineto2025cost}.

The admissible lower-bound tests select $V' = \mathrm{vars}(a)$ and use the same planner
configuration as per the weak tasks.

Experiments were run on a 2022 \textsc{MacBook Pro} with an
\textsc{Apple M2} processor, 24~GB RAM, and
\textsc{Sequoia 15.5}. Unless specified otherwise, reported runtimes
exclude compilation time. Additionally, they are aggregated per candidate action: a weak or strong
run includes all per-effect tasks constructed for that action. Thus the reported run counts refer to actions analysed, not individual calls to the underlying planners.

Table~\ref{tab:classification} reports grounded-action counts, counts
passing the structural precheck, restricted weak and strong
certificates, \textsc{NoStrong} ($\neg\mathsf{Str}$), merge failures, certified
irreversibility, and unclassified actions. Timeouts are zero everywhere, so we do not report them for space constraints. Table~\ref{tab:budgets}
reports the minimum, quartiles, mean, maximum, and standard deviation
of certified finite budgets. The additional tables in this appendix report
per-domain budgets, sensitivity ratios, lower-bound rejection, and
runtime.

\subsection{Per-domain reversal budgets}

Table~\ref{tab:budgets-domain} reports finite restricted budgets by
domain under unit costs.

\begin{table*}[t]
\centering
\small
\begin{tabular}{lrrrrrr}
\toprule
Domain
& $\#\mathsf{Wk}^{\mathsf{u}}_{\mathrm{res}}$
& Mean $k_{\mathrm{wk}}$
& Max $k_{\mathrm{wk}}$
& $\#\mathsf{BStr}^{\mathsf{u}}_{\mathrm{res}}$
& Mean $k_{\mathrm{str}}$
& Max $k_{\mathrm{str}}$\\
\midrule
\begin{filecontents*}{results/table3_unit.csv}
Domain,nWkRes,MeanWk,MaxWk,nBStrRes,MeanStr,MaxStr
acrobatics,4,1.5,3,3,1,1
beam-walk,6,4,6,0,-,-
blocksworld,5,1,1,5,1,1
bus-fare,2,1,1,0,-,-
climber,0,-,-,0,-,-
doors,0,-,-,0,-,-
earth-observation,21,2,2,21,2,2
elevators,8,1,1,8,1,1
ex-blocksworld,0,-,-,0,-,-
faults,0,-,-,0,-,-
first-responders,28,0.786,1,28,0.786,1
forest,10,0.4,1,6,0,0
islands,20,1,1,20,1,1
miner,158,1,1,158,1,1
river,0,-,-,0,-,-
spiky-tireworld,124,1,1,124,1,1
tireworld,0,-,-,0,-,-
tireworld-truck,10,1,1,10,1,1
triangle-tireworld,0,-,-,0,-,-
zenotravel,504,7.19,9,0,-,-
\end{filecontents*}
\csvreader[
  head to column names,
  late after line=\\
]{results/table3_unit.csv}{}{%
\textsc{\Domain}
& \nWkRes
& \MeanWk
& \MaxWk
& \nBStrRes
& \MeanStr
& \MaxStr
}%
\bottomrule
\end{tabular}
\caption{Per-domain finite restricted budget statistics under unit costs.
The weak columns summarize $k^{\star,\mathsf{u}}_{\mathrm{wk,res}}$;
the strong columns summarize
$k^{\star,\mathsf{u}}_{\mathrm{str,res}}$.}
\label{tab:budgets-domain}
\end{table*}

The aggregate weak-budget distribution is strongly influenced by
\textsc{zenotravel}: it contributes $504$ weakly certified actions
with mean weak budget $7.19$ and maximum $9$, but no finite restricted
strong budgets. By contrast, \textsc{miner},
\textsc{spiky-tireworld}, and \textsc{islands} contain many actions
whose weak and finite strong budgets are both $1$.

The unit-cost classification also varies by domain. For example,
\textsc{earth-observation}, \textsc{miner}, \textsc{islands},
\textsc{spiky-tireworld}, and \textsc{tireworld-truck} have many
finite restricted strong witnesses, whereas the $504$ weakly certified
\textsc{zenotravel} actions have constructed restricted strong-cyclic
witnesses. The $\textsc{?}$ column in
Table~\ref{tab:classification} is not a negative classification:
it records actions for which the sufficient tests establish none of
the reported classes.

\paragraph{Comparison with~\citet{chrpa2024weak}.} Our cost-sensitive results refine, rather than contradict, qualitative strong reversibility: the 504 \textsc{zenotravel} actions reported as strongly universally uniformly reversible by Med et al. have constructed strong-cyclic witnesses in our restricted compilation, guaranteeing eventual reversal under fairness but not a finite worst-case cost. Med et al.’s published erratum also corrects the strong classifications for \textsc{beam-walk} and \textsc{forest} to weak-only; our results should be compared against those corrected classifications. We also corrected the handling of empty goals, which can arise from actions with empty preconditions. The earlier pipeline classified these cases as irreversible rather than unclassified (?), although the empty goal is satisfied by the empty plan and therefore cannot certify irreversibility. This correction changes the \textsc{forest} “?” count from 129 to 130 in our analysis and \textsc{I} from 17 to 16 (see their second minor erratum for further details). Our grounded-action totals differ for \textsc{first-responders} (52 versus 46) and \textsc{forest} (156 versus 150) because we removed the no-op filter used in Med et al.’s determinized SAS+ representation. That filter discarded entirely no-op grounded actions, such as \texttt{move-plane a1 l21 l21} and \texttt{drive-fire-unit f1 l3 l3}, accounting for the action-count differences. More importantly, it also discarded no-change determinizations corresponding to empty branches of \texttt{oneof} effects. We retain those branches because they are genuine outcomes; omitting them changes the transitions considered in reversal, particularly when assessing whether recovery can cycle. Accordingly, we compare classifications on matching grounded actions rather than treating the two tables’ totals as directly comparable.

\subsection{Sensitivity to the cost model}

For actions certified under multiple cost models, we compare finite
budgets through the ratios
$$
  r_{\mathrm{eff/unit}}(a)
  =
  \frac{k_{\mathrm{eff}}(a)}{k_{\mathrm{unit}}(a)}
  \quad\text{and}\quad
  r_{\mathrm{nd/unit}}(a)
  =
  \frac{k_{\mathrm{nd}}(a)}{k_{\mathrm{unit}}(a)},
$$
computed separately for weak and strong budgets. Table~\ref{tab:cost-sensitivity}
summarizes the results.

\begin{table*}[t]
\centering
\small
\begin{tabular}{llrrrrrr}
\toprule
Budget type
& Ratio
& $\#$
& Min
& Median
& Mean
& Max
& Std.\\
\midrule
\begin{filecontents*}{results/table4.csv}
BudgetType,Ratio,N,Min,Median,Mean,Max,Std
weak,eff/unit,888,2.0,3.889,3.66,4.0,0.514
weak,nd/unit,888,2.0,4.0,4.002,5.0,0.675
strong,eff/unit,371,2.0,4.0,3.426,4.0,0.654
strong,nd/unit,371,2.0,4.0,3.426,4.0,0.654
\end{filecontents*}
\csvreader[
  head to column names,
  late after line=\\
]{results/table4.csv}{}{%
\ifcsvstrcmp{\BudgetType}{weak}{Weak}{Strong}
&
\ifcsvstrcmp{\Ratio}{eff/unit}
  {$r_{\mathrm{eff/unit}}$}
  {$r_{\mathrm{nd/unit}}$}
& \N
& \Min
& \Median
& \Mean
& \Max
& \Std
}%
\bottomrule
\end{tabular}
\caption{Sensitivity of finite reversal budgets to the action-cost model.
Ratios are computed only for actions certified under both compared cost
models and with finite budgets.}
\label{tab:cost-sensitivity}
\end{table*}

The qualitative classification is stable across these cost models,
but numerical budgets are not. Relative to unit costs, finite weak
budgets increase by a mean factor of $3.66$ under effect-size costs
and about $4.00$ under outcome-sensitive costs. Finite strong budgets
increase by about $3.43$ under either non-unit model. For the finite
strong witnesses found here, the selected reversing actions incur
no additional outcome-sensitive penalty beyond the effect-size
component.

\subsection{Lower-bound rejection}

We test the admissible rejection procedure of
Theorem~\ref{thm:irrcomp} at
$K=\{1,3,5,10,15\}$ under each cost model. For each action and cost model, we compute the lower-bound cost
once and reuse it to test all five thresholds in $K$. Table~\ref{tab:lower-bound} reports the results.
For each action rejected at $k\in K$, we compare $k$ with its restricted
weak budget $k^{\star,u}_{\mathrm{wk,res}}(a)$ from Theorem~4.
We report a contradiction if
$k^{\star,u}_{\mathrm{wk,res}}(a) \leq k$, since the restricted
policy then witnesses full-domain reversal within $k$; when
$k^{\star,u}_{\mathrm{wk,res}}(a) > k$, the two procedures are
consistent (as they are when a per-effect task is unsolvable), but this comparison alone does not independently
establish the full-domain optimum. Finally, if the test does not reject an action at budget $k$,
its result for that action and budget is inconclusive.
\begin{table*}[t]
\centering
\small
\begin{tabular}{llrrrrrr}
\toprule
Cost model
& $k$
& Tested actions
& Rejected
& Rejection rate
& Consistent rejections
& Inconsistencies
& Inconclusive\\
\midrule
\begin{filecontents*}{results/table5.csv}
CostModel,k,TestedActions,Rejected,RejectionRate,TruePositives,FalsePositives,Inconclusive
unit,1,1178,788,0.669,788,0,390
unit,3,1178,524,0.445,524,0,654
unit,5,1178,486,0.413,486,0,692
unit,10,1178,64,0.054,64,0,1114
unit,15,1178,64,0.054,64,0,1114
effect,1,1178,1166,0.99,1166,0,12
effect,3,1178,993,0.843,993,0,185
effect,5,1178,767,0.651,767,0,411
effect,10,1178,656,0.557,656,0,522
effect,15,1178,522,0.443,522,0,656
outcome,1,1178,1166,0.99,1166,0,12
outcome,3,1178,997,0.846,997,0,181
outcome,5,1178,775,0.658,775,0,403
outcome,10,1178,704,0.598,704,0,474
outcome,15,1178,606,0.514,606,0,572
\end{filecontents*}
\csvreader[
  head to column names,
  late after line=\\
]{results/table5.csv}{}{%
\ifcsvstrcmp{\CostModel}{unit}{Unit}{%
  \ifcsvstrcmp{\CostModel}{effect}{Effect-size}{Outcome-sensitive}%
}
& \k
& \TestedActions
& \Rejected
& \RejectionRate
& \TruePositives
& \FalsePositives
& \Inconclusive
}%
\bottomrule
\end{tabular}
\caption{Results and effectiveness of the lower-bound rejection test by cost model and budget $k$.
A rejection is consistent when a finite restricted weak budget
$k^{\star,u}_{\mathrm{wk,res}}(a)$ exceeds $k$, or when at least
one restricted per-effect task is unsolvable. An inconsistency
occurs when $k^{\star,u}_{\mathrm{wk,res}}(a)\leq k$, since the
restricted policy then witnesses reversal within $k$.
Inconclusive counts actions not rejected by the lower-bound test.
Consistency does not establish the full-domain optimum.}
\label{tab:lower-bound}
\end{table*}

Under unit costs, the test rejects $788$ of $1178$ structurally
eligible actions at $k=1$ ($66.9\%$), falling to $64$ at
$k=10$ and $k=15$. At $k=1$, both non-unit models reject
$1166$ actions ($99.0\%$). No inconsistencies were observed. A non-rejection does not certify a
$k$-bounded witness; the test is a negative filter, not a
replacement for the optimal compilations.

\subsection{Runtime and scalability}

Table~\ref{tab:runtime} reports aggregate runtimes, and
Table~\ref{tab:runtime-domain} reports per-domain means.

\begin{table*}[t]
\centering
\small
\begin{tabular}{lrrrrrr}
\toprule
Procedure
& Action analysed
& Solved
& Timeouts
& Solve mean
& Solve max
& Total mean\\
\midrule
\begin{filecontents*}{results/table6_unit.csv}
Procedure,Calls,Solved,Timeouts,CompMean,CompMax,SolveMean,SolveMax,TotalMean
Weak classical compilation,1178,1178,0,-,-,45.576,170.079,45.576
Strong FOND compilation,900,900,0,-,-,4179.651,6959.238,4179.651
Irreversibility compilation,971,971,0,-,-,65.25,195.146,65.25
\end{filecontents*}
\csvreader[
  head to column names,
  late after line=\\
]{results/table6_unit.csv}{}{%
\Procedure
& \Calls
& \Solved
& \Timeouts
& \SolveMean
& \SolveMax
& \TotalMean
}%
\bottomrule
\end{tabular}
\caption{Runtime statistics in milliseconds unless otherwise stated. Solver time is aggregated over the per-effect tasks for each action; means and maxima are over actions.}
\label{tab:runtime}
\end{table*}

All planner calls terminate within the timeout. Mean solve times are
$45.576$ ms over $1178$ actions analysed for weak reversal, $65.250$ ms over $971$ action analysed for
irreversibility, and $4179.651$ ms over $900$ actions analysed for strong reversal. Strong search is significantly more expensive because it leverages a planner whose implementation is not efficiency oriented.

\begin{table*}[t]
\centering
\small
\begin{tabular}{lrrrrrr}
\toprule
Domain
& Weak actions
& Weak mean
& Strong actions
& Strong mean
& LB actions
& LB mean\\
\midrule
\begin{filecontents*}{results/table7_unit.csv}
Domain,WeakCalls,WeakMean,StrongCalls,StrongMean,LBCalls,LBMean
acrobatics,5,51.662,4,1142.97,5,46.268
beam-walk,7,57.904,6,1123.736,7,55.339
blocksworld,5,81.229,5,1657.843,5,85.614
bus-fare,5,61.054,2,1683.78,5,70.986
climber,3,40.859,0,-,3,52.772
doors,0,-,0,-,0,-
earth-observation,27,44.341,21,971.438,27,53.997
elevators,8,44.114,8,860.924,8,42.772
ex-blocksworld,5,41.726,0,-,5,43.689
faults,0,-,0,-,0,-
first-responders,42,43.683,28,843.001,42,45.684
forest,30,56.803,10,1616.44,30,97.482
islands,24,48.292,20,1010.714,24,46.273
miner,176,42.019,158,1794.443,176,41.652
river,1,41.102,0,-,1,78.101
spiky-tireworld,130,43.7,124,1879.839,130,41.492
tireworld,44,81.104,0,-,44,119.03
tireworld-truck,14,41.195,10,866.772,14,39.545
triangle-tireworld,8,40.131,0,-,8,78.433
zenotravel,644,43.712,504,6202.688,644,48.441
\end{filecontents*}
\csvreader[
  head to column names,
  late after line=\\
]{results/table7_unit.csv}{}{%
\textsc{\Domain}
& \WeakCalls
& \WeakMean
& \StrongCalls
& \StrongMean
& \LBCalls
& \LBMean
}%
\bottomrule
\end{tabular}
\caption{Per-domain runtime statistics. Mean runtimes are computed over actions, not individual planner
calls. LB time includes one lower-bound cost computation per action
and its evaluation against all five budget thresholds.}
\label{tab:runtime-domain}
\end{table*}

\textsc{zenotravel} has the highest reported mean strong solve time:
$6202.688$ ms over $504$ calls. The corresponding means are
$1794.443$ ms for \textsc{miner} and $1879.839$ ms for
\textsc{spiky-tireworld}. Lower-bound-test means are typically below
$100$ ms per domain, supporting its use as a preprocessing filter
when the proposed budget is small.

\subsection{Limitations}

\paragraph{Restricted versus full-domain budgets.}
The finite weak and strong budgets are exact for the restricted
compilations but only upper bounds on full-domain optima. An
unrestricted policy may find a cheaper reversal using variables or
actions outside $\mathrm{vars}(\mathrm{pre}(a))$.

\paragraph{Restricted strong cyclicity and incompleteness.}
A constructed restricted strong-cyclic witness has unbounded
worst-case finite cost in the restricted search space; a full-domain
acyclic witness may nevertheless exist. The pipeline supplies sound
but incomplete certificates. In particular, a \textsc{MergeFail}
result does not prove that no strong uniform policy exists, and
failure of the lower-bound test to reject at $k$ does not prove
$k$-bounded reversibility.

\paragraph{Synthetic costs.}
The three reproducible cost models are not domain-specific utility
models. In an application, action costs should reflect the relevant
recovery burden, such as elapsed time, resources, risk, or user effort.

\paragraph{Relational policies.}
A weak witness may assign multiple actions to one state. Turning it
into a deterministic controller may require selecting an action with
the current reversal target as context. That execution issue is
separate from the existence and budget certificates evaluated here.

\end{document}